\documentclass[sigconf]{acmart}
\AtBeginDocument{%
  }

\usepackage{subcaption}
\usepackage{algorithm}
\usepackage{algpseudocode}

\DeclareMathOperator*{\argmax}{argmax} % thin space, limits underneath in displays
\usepackage[official]{eurosym}

\copyrightyear{2024} 
\acmYear{2024} 
\setcopyright{rightsretained} 
\acmConference[HAI '24]{International Conference on Human-Agent Interaction}{November 24--27, 2024}{Swansea, United Kingdom}
\acmBooktitle{International Conference on Human-Agent Interaction (HAI '24), November 24--27, 2024, Swansea, United Kingdom}\acmDOI{10.1145/3687272.3688298}
\acmISBN{979-8-4007-1178-7/24/11}

\begin{document}

%%
%% The "title" command has an optional parameter,
%% allowing the author to define a "short title" to be used in page headers.
\title{``Give Me an Example Like This'': Episodic Active Reinforcement Learning from Demonstrations}

%%
%% The "author" command and its associated commands are used to define
%% the authors and their affiliations.
%% Of note is the shared affiliation of the first two authors, and the
%% "authornote" and "authornotemark" commands
%% used to denote shared contribution to the research.
\author{Muhan Hou}
\email{m.hou@vu.nl}
\affiliation{%
  \institution{Vrije Universiteit Amsterdam}
  \city{Amsterdam}
  \country{the Netherlands}
}

\author{Koen Hindriks}
\email{k.v.hindriks@vu.nl}
\affiliation{%
  \institution{Vrije Universiteit Amsterdam}
  \city{Amsterdam}
  \country{the Netherlands}
}

\author{A.E. Eiben}
\email{a.e.eiben@vu.nl}
\affiliation{%
  \institution{Vrije Universiteit Amsterdam}
  \city{Amsterdam}
  \country{the Netherlands}
}

\author{Kim Baraka}
\email{k.baraka@vu.nl}
\affiliation{%
  \institution{Vrije Universiteit Amsterdam}
  \city{Amsterdam}
  \country{the Netherlands}
}

%%
%% By default, the full list of authors will be used in the page
%% headers. Often, this list is too long, and will overlap
%% other information printed in the page headers. This command allows
%% the author to define a more concise list
%% of authors' names for this purpose.
\renewcommand{\shortauthors}{Hou M., et al.}

%%
%% The abstract is a short summary of the work to be presented in the
%% article.
\begin{abstract}
 Reinforcement Learning (RL) has achieved great success in sequential decision-making problems but often requires extensive agent-environment interactions. To improve sample efficiency, methods like Reinforcement Learning from Expert Demonstrations (RLED) incorporate external expert demonstrations to aid agent exploration during the learning process. However, these demonstrations, typically collected from human users, are costly and thus often limited in quantity. Therefore, how to select the optimal set of human demonstrations that most effectively aids learning becomes a critical concern. This paper introduces EARLY (Episodic Active Learning from demonstration querY), an algorithm designed to enable a learning agent to generate optimized queries for expert demonstrations in a trajectory-based feature space. EARLY employs a trajectory-level estimate of uncertainty in the agent's current policy to determine the optimal \textit{timing} and \textit{content} for feature-based queries. By querying episodic demonstrations instead of isolated state-action pairs, EARLY enhances the human teaching experience and achieves better learning performance. We validate the effectiveness of our method across three simulated navigation tasks of increasing difficulty. Results indicate that our method achieves expert-level performance in all three tasks, converging over $\mathbf{50}$\textbf{\%} faster than other four baseline methods when demonstrations are generated by simulated oracle policies. A follow-up pilot user study ($N=18$) further supports that our method maintains significantly better convergence with human expert demonstrators, while also providing a better user experience in terms of perceived task load and requiring significantly less human time.
\end{abstract}

%%
%% The code below is generated by the tool at http://dl.acm.org/ccs.cfm.
%% Please copy and paste the code instead of the example below.
%%



%%
%% Keywords. The author(s) should pick words that accurately describe
%% the work being presented. Separate the keywords with commas.
%% A "teaser" image appears between the author and affiliation
%% information and the body of the document, and typically spans the
%% page.
% \begin{teaserfigure}
%   \includegraphics[width=\textwidth]{sampleteaser}
%   \caption{Seattle Mariners at Spring Training, 2010.}
%   \Description{Enjoying the baseball game from the third-base
%   seats. Ichiro Suzuki preparing to bat.}
%   \label{fig:teaser}
% \end{teaserfigure}

% \received{20 February 2007}
% \received[revised]{12 March 2009}
% \received[accepted]{5 June 2009}

%%
%% This command processes the author and affiliation and title
%% information and builds the first part of the formatted document.
\maketitle

%%%%%%%%%%%%%%%%%%%%%%%%%%%%%%%%%%%%%%%%%%%%%%%%%%%%%%%%%%%%%%%%%%%%%%%%

\section{Introduction}

\begin{figure*}[t!]
  \centering
  \includegraphics[width=0.7\linewidth]{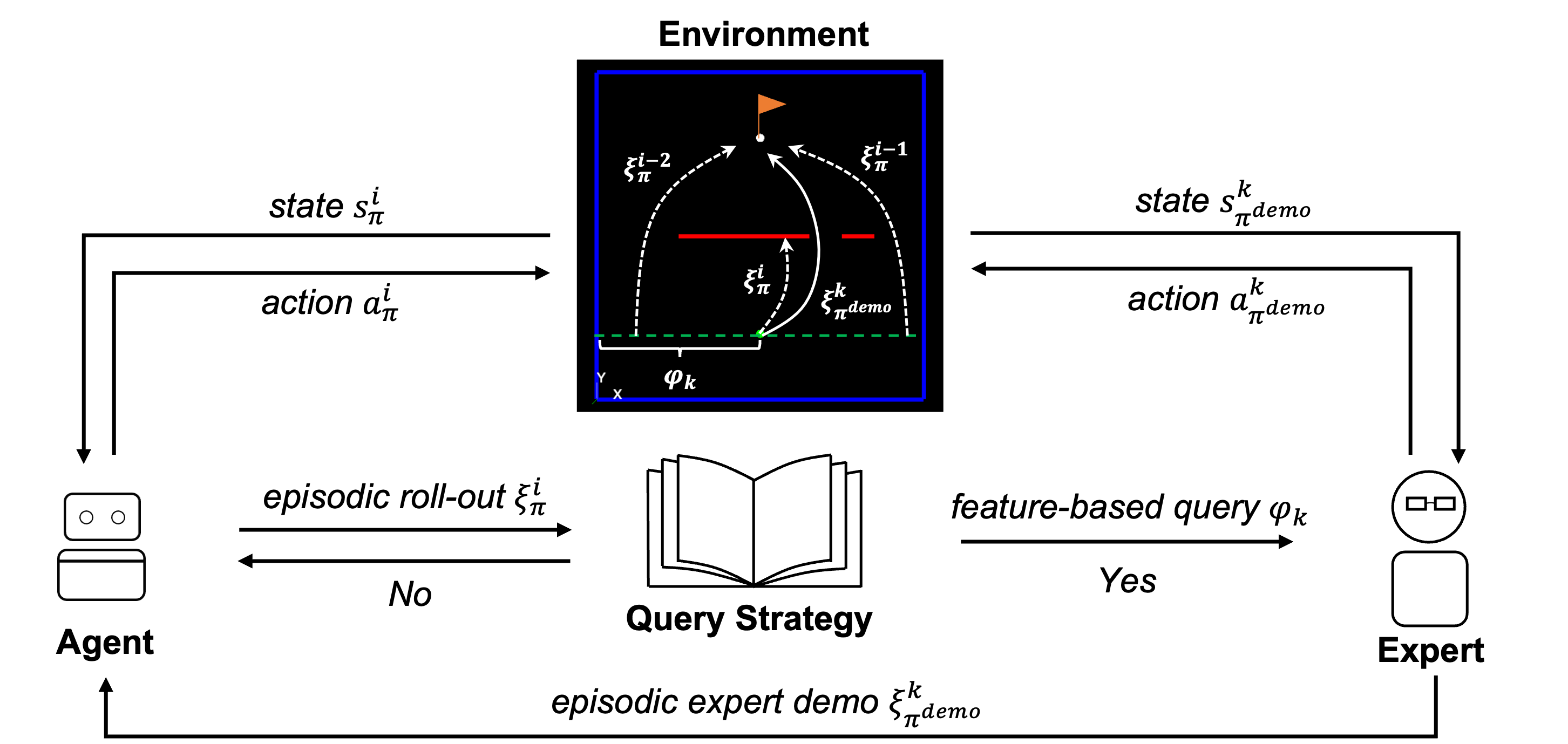}
  \caption{Overview of our method. After each of the episodic roll-out $\xi_{\pi}^i$, our query strategy will evaluate the uncertainty of $\xi_{\pi}^i$ based on a trajectory-based uncertainty measurement, and determine whether and what to query via a dynamic adaptive threshold for uncertainty. Once to query, a feature-based query $\varphi_k$ will be made for an episodic expert demonstration $\xi_{\pi^{demo}}^k$, whose feature value is expected to be of the queried $\varphi_k$ (e.g., ``give me a demonstration that starts from this initial position and arrive at the destination'' when the feature is defined as the initial state of a roll-out trajectory). This process will continue until all expert demonstrations are collected.}
  \label{overview}
\end{figure*}

Reinforcement Learning (RL) is a widely used approach for tackling problems involving sequential decision-making. In RL, an agent learns to improve its policy through trial-and-error interactions with the environment, aiming to maximize the expected long-term rewards. However, this approach often requires extensive agent-environment interactions before achieving a high-quality policy. To enhance sample efficiency, methods such as Reinforcement Learning from Expert Demonstrations (RLED) ~\cite{ramirez2022model} leverage expert demonstrations to accelerate the learning process. By adopting a \textit{demo-then-training} strategy, these methods significantly reduce the required interactions, enabling the agent's policy to converge to an expert-level policy much faster ~\cite{vecerik2017leveraging, nair2018overcoming, kang2018policy}.

Despite the advantages that expert demonstrations may provide, collecting them can be time-consuming and expensive, especially when they come from real human experts. In practice, the number of demonstrations is typically constrained by a limited budget. Therefore, \textit{how to select the optimal set of demonstrations} to maximize their benefit to agent learning becomes a crucial consideration.

However, selecting the distribution of demonstrations is intertwined with the policy learning process itself, making it challenging to determine which distribution would be most beneficial before learning begins. In the case of human experts, even if a human expert could demonstrate the optimal action to take for every state encountered in any chosen demonstration (i.e., \textit{optimal in executing the task}), the overall distribution of selected demonstrations itself might not be optimal for learning (i.e., \textit{sub-optimal in teaching the task}). One intuitive strategy is to cover as many diverse areas of state space with demonstrations as possible. However, without proper guidance, the natural distribution of collected demonstrations often results in an uneven coverage of the state space ~\cite{hou2023shaping}. Moreover, such a uniform coverage strategy is not necessarily optimal for policy learning. For critical areas of the state space that might be less frequent to encounter but much harder for the control policy to be generalized to (e.g., encountering an oncoming vehicle in an autonomous driving setting), they might require more expert demonstrations than those that are more frequent to encounter but much easier to handle (e.g., driving straight when there are no vehicles around) ~\cite{hawke2020urban}. Defining these critical situations is often task-specific and influenced by inherent differences in cognitive patterns between human experts and algorithm-driven agents. Situations that human experts perceive as easy to learn may prove difficult for learning agents to generalize, and vice versa. Moreover, the probability distribution of running into different areas of state space is non-stationary during the learning process, as it depends on the evolving agent policy that iteratively updates its action distributions over states. This dynamic nature makes it even more impractical to determine the optimal distribution of demonstrations before policy learning begins.

Alternatively, efforts have also been made to let agents learn in a \textit{demo-while-training} manner and actively request teaching inputs that are most beneficial for them during the learning process. A common paradigm for these methods is to measure the informativeness (e.g., uncertainty ~\cite{chen2020active, chenactive}, discrepancy ~\cite{subramanian2016exploration}, etc.) of each encountered state as the learning agent rolls out its current policy, switch or share the control with an expert demonstrator at certain threshold, and let the agent regain full autonomy when it is back to normal. However, such a paradigm tends to be time-consuming. Since each control switch requires the task environment to be reset to several moments prior for context, it will inevitably consume much more human time \cite{johns2022back} due to these contextual replays. Furthermore, it is cognitively demanding and susceptible to noises, particularly in real-world scenarios where environment resetting is impractical. In such cases, human experts have to be fully engaged throughout the learning process and ready for immediate intervention that may be requested at any time. This will pose a great cognitive load on human demonstrators and can easily introduce noise or errors in providing immediate intervention ~\cite{kelly2019hg}.

% Furthermore, human decision-making relies not only on the observation of the current state and future expectations but also on the near history that leads to the current observation. Therefore, when presenting the isolated states for human demonstrations, it often requires rewinding the environment to the past few steps to facilitate demonstrators. However, such a replaying demand in interface might simply be impractical to realize when it comes to embodied learning agents (e.g., to rewind a moving ball to its previous position and velocity in order to demonstrate to a real robot how it should have hit the ball).

% In the example of autonomous navigation, instead of asking for immediate expert actions from the human expert whenever the agent runs into an uncertain situation, our method will query a complete episodic expert demonstration of certain target feature values, such as a demonstration that starts from a certain initial position, arrives at a certain destination, or reaches a maximum velocity of a certain value, etc.

To alleviate the demanding cognitive loads and overcome the disturbance issues caused by isolated state-based queries, we present a method that enables an RL agent to actively request episodic demonstrations (i.e., starting from an initial state till a terminal state) for better learning performance and improved user experience, as shown in Figure ~\ref{overview}. To achieve these, we construct a trajectory-based uncertainty measurement to evaluate episodic policy roll-outs and utilize it to optimize the decision of \textit{when to query} and \textit{what to query} in a trajectory-based feature space. We test our method on three simulated navigation tasks with sparse rewards, a continuous state action space, and increasing levels of difficulty. Compared with $4$ other popular baselines, our results indicate that our method converges to expert-level performance significantly faster in both experiments with oracle-simulated demonstrators and real human expert demonstrators while achieving improved perceived task load and consuming significantly less human time. 

In summary, our main contributions are as follows:
\begin{itemize}
    \item We design EARLY, an episode-based query algorithm that is built in trajectory-based feature space to actively determine \textit{when} to query and \textit{what} episodic expert demonstration to query.

    \item We propose a trajectory-based uncertainty measurement of the agent policy based on temporal difference errors of episodic policy roll-out.
    
    \item We validate the effectiveness of our method in learning performance and user experience with both simulated oracle and real human expert demonstrators.
\end{itemize}

%%%%%%%%%%%%%%%%%%%%%%%%%%%%%%%%%%%%%%%%%%%%%%%%%%%%%%%%%%%%%%%%%%%%%%%%

\section{Related Work}
To improve the sample efficiency of conventional RL methods, much effort has been made to introduce teaching input into the learning loop. These external inputs (e.g., demonstrations) are either passively or actively utilized by the learning agent, aiming to guide the policy exploration and accelerate the training process.
\subsection{Reinforcement Learning from Demonstrations}
Deep Q-Learning from Demonstrations (DQfD) ~\cite{hester2018deep} leverages expert demonstrations to accelerate off-policy training. By adding demonstrations to the reply buffer of Deep Q-Learning (DQN) ~\cite{mnih2013playing}, it greatly facilitates the policy exploration for tasks of a discrete action space. Deep Deterministic Policy Gradient from Demonstrations (DDPGfD) ~\cite{vecerik2017leveraging} extends DQfD to tasks with a continuous action space and sparse rewards. It introduces an n-step return loss to more accurately estimate the temporal difference error and uses the reply buffer with Prioritized Experience Replay (PER) ~\cite{schaul2015prioritized} to better balance the sampling between agent roll-outs and expert demonstrations. Nair et al. ~\cite{nair2018overcoming} further improved the applicability of DDPGfD to more complicated robotic tasks. Policy Optimization from Demonstration (POfD) ~\cite{kang2018policy} also leverages demonstrations to guide policy exploration, and it employs the occupancy measure to make the algorithm less susceptible to the amount limitation and sub-optimality of demonstrations. Other works further extend the usage of demonstrations to various task settings ~\cite{taylor2011integrating, wang2017improving, singh2020parrot, nair2020awac} and real-world applications ~\cite{liu2022improved}.

\subsection{Active Learning from Demonstrations}
Instead of passively receiving demonstrations and updating the policy based on them, recent work attempted to enable the learning agent to learn in a \textit{demo-while-training} manner and actively request demonstrations, which may alleviate the issue of covariance shift and accelerate the learning process. For instance, Confidence-Based Autonomy (CBA) ~\cite{chernova2009interactive} estimates the state uncertainty based on the classification confidence of agent actions in the setting of supervised learning. The agent will query a demonstration for the current state when its uncertainty exceeds a threshold that is determined by the classifier decision boundary. Subramanian et al. ~\cite{subramanian2016exploration} evaluate the state uncertainty with statistical measures called leverage and discrepancy to find important states and query demonstrations that are able to reach these states to guide policy exploration. Selective Active Learning from Traces (SALT) ~\cite{packard2017policies} constructs a query strategy based on accumulated rewards and request demonstrations when the encountered state is quite different from the already collected roll-out steps. Active Reinforcement Learning with Demonstrations (ARLD) ~\cite{chen2020active} estimates the uncertainty of each encountered state via Q-value-based measurements and generates a dynamic adaptive uncertainty threshold to determine the query timing. Chen et al. ~\cite{chenactive} extend ARLD to tasks of continuous action spaces and construct a new uncertainty measurement of individual states based on the variance of actions produced by the agent policy. By contrast, Rigter et al. ~\cite{rigter2020framework} present a framework that generates demonstration queries by explicitly taking into account the human time cost for demonstrating and the risk of agent policy failure. Furthermore, some efforts have also been made to combine active learning with Learning from Demonstrations (LfD) in scenarios where reward signals are not available ~\cite{judah2014active} and multiple query types can be chosen from ~\cite{cakmak2011active}, and to solve real-world tasks ~\cite{silver2012active, hou2023shaping}. However, most of these efforts have been focused on the teaching input of isolated state-action pairs, which have to be requested from demonstrators via frequent contextual switches. Although some work ~\cite{silver2012active} also attemped to actively utilize episodic demonstrations, it reduced the problem into a model-based supervised learning setup with offline datasets. Such an approach may not be valid for sequential decision making where the state distribution is non-stationary. By contrast, our work is focused on using episodic demonstrations for sequential decision-making scenarios and aims to improve user experience while accelerating policy learning at the same time.

%%%%%%%%%%%%%%%%%%%%%%%%%%%%%%%%%%%%%%%%%%%%%%%%%%%%%%%%%%%%%%%%%%%%%%%%

% \subsection{Replay Buffer Design in Off-Policy RL}

%%%%%%%%%%%%%%%%%%%%%%%%%%%%%%%%%%%%%%%%%%%%%%%%%%%%%%%%%%%%%%%%%%%%%%%%

\section{Methodology}
We present a method that enables the learning agent to actively request episodic expert demonstrations that are most beneficial for its learning while optimizing its own policy in an off-policy manner. Similar to ~\cite{chenactive}, we choose Soft Actor-Critic (SAC) ~\cite{haarnoja2018soft} as the underlying off-policy RL algorithm for its superior performance in tasks with continuous state-action spaces. Furthermore, instead of querying isolated state-action pairs in state space as in ~\cite{chen2020active} and ~\cite{chenactive}, we design a query strategy constructed in a trajectory-based feature space where we evaluate policy uncertainty and query episodic expert demonstrations.

\subsection{Problem Setup} \label{problem setup}
We formulate the problem of active learning from demonstrations as a Markov Decision Process (MDP). We assume that the specifications $(S, A, r, \gamma, P)$ of the MDP are given, where $S$ is the state space, $A$ is the action space, $r(s_t, a_t): S \times A \rightarrow \mathbb{R}$ is the reward function, and $\gamma$ is the discount factor. For the transition function $P(s_{t+1} | s_t, a_t)$, we assume that its explicit expression is unknown but a task environment is available for unlimited interactions.

Furthermore, we also assume that episodic demonstrations are available upon querying an expert $\pi_{demo}$, which is optimal or close to optimal. We assume that only a limited number of demonstrations can be provided during the agent learning process, and this amount budget of $N_d$ is known before the learning process starts.

We assume that the feature vector $\varphi_i \in \Phi$ of a policy episodic roll-out trajectory $\xi_{\pi}^i=\{(s_t^i, a_t^i, r_t^i, s_{t+1}^i)\}_{t=0}^{T-1}$ of length $T$ can be obtained via a given feature function $\Phi(\cdot)$ (i.e., $\varphi_i = \Phi(\xi_{\pi}^i)$). Under a policy $\pi_{\phi}$ parametrized by $\phi$, the probability of obtaining the episodic roll-out trajectory $\xi^{i}_{\pi}$ is

\begin{equation}
    P(\xi^i_{\pi}; \phi) = \mu(s_0^i)\prod_{t=0}^{T-1} P( s_{t+1}^i | s_t^i, a_t^i ) \pi_{\phi} (a_t^i | s_t^i),
\end{equation}

where $\mu(s_0^i)$ is the initial state distribution independently determined by the task environment. Therefore, the probability of obtaining a roll-out trajectory whose feature value is of $\varphi_i$ will be

\begin{align}
    P(\varphi_i; \phi) & = \sum_{\xi_{\pi}^j \in D_{\varphi_i}}P(\xi_{\pi}^j; \phi) \\
                             & = \sum_{\xi_{\pi}^j \in D_{\varphi_i}}\mu(s_0^j) \prod_{t=0}^{T-1}P(s_{t+1}^j | s_t^j, a_t^j ) \pi_{\phi} (a_t^j | s_t^j),
\end{align}

where $D_{\varphi_i}$ represents the set of all roll-out trajectories under the current agent policy $\pi$ whose feature values are equal to $\varphi_i$.

By contrast, when the agent generates a feature-based query $\varphi_k$ and queries for an episodic expert demonstration whose feature value is expected to be of $\varphi_k$ (e.g., "Give me an episodic demonstration of this target feature value."), the probability of the agent obtaining such an expert demonstration $\xi_{\pi^{demo}}^i$ is
\begin{equation}
    P(\xi^i_{\pi^{demo}}; \varphi_k) = \mu(s_0^i; \varphi_k) \prod_{t=0}^{T-1} P( s_{t+1}^i | s_t^i, a_t^i ) \pi^{demo} (a_t^i | s_t^i),
\end{equation}
where $\mu(s_0^i; \varphi_k)$ represents the initial state distribution of expert demonstrations that is influenced by the feature-based query $\varphi_k$.

To simplify the problem, in this work, we chose the initial state $s_0$ of a roll-out trajectory as its feature. This will make $P(\varphi_i; \phi)$ only depend on the initial state distribution $\mu(s_0)$ and not affected by the current policy $\pi$. Furthermore, when the agent queries an episodic demonstration from the expert, we assume that the agent will always be able to get an expert demonstration whose feature value is exactly of the queried feature value (i.e., starting from the queried initial state), leading to $P(\xi^i_{\pi^{demo}}; \varphi_k)=\mu(s_0^i; \varphi_k)=\delta(\varphi_k)$, where $\delta(\cdot)$ represents the Dirac delta distribution. 

By actively generating feature-based queries and asking for corresponding episodic expert demonstrations, the goal of our method is to design a query strategy to wisely determine when to query and what to query so as to make the most of a limited number of queries and help the agent policy approximate expert policy with as few environment interactions as possible.

% Furthermore, we also assume that an inverse model $\Phi^{-1}_{\pi}(\cdot)$ is given so that an episodic roll-out trajectory $\xi_{\pi}$, whose feature value is of some target feature value $\varphi \in \Phi$, can be generated under a target policy $\pi$ (i.e., $\xi_{\pi} = \Phi^{-1}_{\pi}(\varphi | \pi)$). Instead of querying isolated states in the state space, we assume each time the agent queries for an episodic expert demonstration that starts from some initial state until reaching a terminal state. Such queries are made by the learning agent to the expert in the feature space (e.g., "Give me an episodic demonstration of this target feature value.").

\subsection{Background on Soft Actor-Critic}

This work builds on Soft Actor-Critic (SAC) ~\cite{haarnoja2018soft}, a state-of-the-art off-policy RL algorithm that employs the actor-critic structure, including a parametrized state-action value function $Q_{\theta}(s_t, a_t)$, a state value function $V_{\psi}(s_t)$, and a stochastic policy $\pi_{\phi}(s_t | a_t)$. To better stabilize training, SAC also includes a parametrized target value function $V_{\bar{\psi}}(s_t, a_t)$ that updates much slower than $V_{\psi}(s_t)$. Similar to other off-policy RL algorithms, it also has a reply buffer $D$ used to store the roll-out data produced by its behavior policy and to be sampled from for updating value functions and policy nets.

During each training iteration, the state value function $V_{\psi}(s_t)$ is updated by minimizing its corresponding cost function $J_V(\psi)$ defined as:
\begin{equation}
    J_V(\psi) = \mathbb{E}_{s_t \sim D}\left[ \frac{1}{2} \left( V_{\psi}\left( s_t \right) - \mathbb{E}_{\pi_{\phi}}\left[Q_{\theta}\left(s_t, a_t \right) - \log\pi_{\phi}\left( a_t | s_t \right)\right] \right)^2  \right].
\end{equation}
To update the state-action value function  $Q_{\theta}(s_t, a_t)$, parameters are optimized by minimizing the cost function $J_Q(\theta)$ defined as:
\begin{equation}
    J_Q(\theta) = \mathbb{E}_{(s_t, a_t) \sim D} \left[ \frac{1}{2} \left( \hat{Q} \left( s_t, a_t \right) - Q_{\theta}\left( s_t, a_t \right) \right)^2 \right],
\end{equation}
where $\hat{Q}(s_t, a_t) = r(s_t, a_t) + \gamma\mathbb{E}_{s_{t+1} \sim p} [ V_{\bar{\psi}}\left( s_{t+1}  \right) ]$ is the target state-action function.
Lastly, the policy net $\pi_{\phi}(s_t|a_t)$ is updated by minimizing
\begin{equation}
    J_{\pi}(\phi) = \mathbb{E}_{s_t \sim D, \epsilon_t \sim \mathcal{N}} \left[ \log\pi_{\phi} \left( f_{\phi}\left( \epsilon_t ; s_t\right) | s_t \right) - Q_{\theta}\left( s_t, f_{\phi} \left( \epsilon_t; s_t \right) \right) \right],
\end{equation}
where $\epsilon_t$ is a noise signal sampled from a given Normal distribution and reparametrized into the original policy net via the transformation $f_{\phi}$ such that $a_t = f_{\phi}(\epsilon_t; s_t)$, aiming to facilitate policy exploration.

\subsection{Trajectory-Based Uncertainty Measurement} \label{uncertainty measurement}
Inspired by ~\cite{gehring2013smart}, we construct an uncertainty measurement for an episodic policy roll-out based on the temporal-difference error. For a given episodic roll-out trajectory $\xi_{\pi}^i$ under the policy $\pi$, we define its uncertainty $u$ as:
\begin{equation}
    u(\xi_{\pi}^i) = \mathbb{E}_{(s_t^i, a_t^i) \in \xi_{\pi}^i} \left[ |\delta_{t}^i| \right],
\end{equation}
with $\delta_t^i$ denoting the temporal-difference error for step $t$ expressed as:
\begin{equation}
    \delta_t^i = r_{t}^i + Q_{\pi}(s_{t+1}^i, a_{t+1}^i) - Q_{\pi}(s_t^i, a_t^i).
\end{equation}
As the absolute value of the temporal-difference error indicates the discrepancy between the target state value and the predicted state value, a higher expectation value of $|\delta_t^i|$ across the state-action pairs along the policy roll-out trajectory intuitively suggests a higher uncertainty of the current policy about this roll-out. Consequently, by querying expert demonstrations that are of the same feature values as those of uncertain roll-outs by the learning agent policy, it may potentially decrease the uncertainties of the areas in the feature space that are around the queried feature points.

\subsection{Episodic Active Reinforcement Learning from Demonstration Query (EARLY)}
Utilizing the trajectory-based uncertainty measurement in Section \ref{uncertainty measurement} and the trajectory-based feature space introduced in Section \ref{problem setup}, we construct an active query strategy for episodic expert demonstrations to solve the problems of \textit{when to query} and \textit{what to query}.

During each training iteration, we first sample an initial state $s_0^i$, obtain an episodic roll-out trajectory $\xi_{\pi}^i$ by the current agent policy $\pi$, and calculate its corresponding feature value $\varphi_i$. To evaluate how the learning agent is uncertain for this feature point, we estimate the uncertainty $u_i$ of the obtained feature point $\varphi_i$ as the agent uncertainty along this generated roll-out trajectory $\xi_{i}^{\pi}$ (i.e., $u_i=u(\xi_{i}^{\pi})$). Both the sampled feature point $\varphi_i$ and its corresponding uncertainty estimation $u_i$ will be stored in shifting recent histories, one for feature points and one for uncertainty values. After the shifting recent history grows to its full length $N_h$, an adaptive uncertainty threshold will be determined via a ratio threshold $r_{query} \in [0, 1]$ as in ~\cite{chen2020active}. Whenever the current uncertainty value $u_i$ is among the top $r_{query}$ of the shifting recent history of uncertainty and the demonstration query budget $N_d$ has not been used up, the learning agent will decide to make a query for one episodic expert demonstration. 

Different from ~\cite{chen2020active}, we choose to query the most uncertain feature point $\varphi_{query}$ in the shifting recent history and ask for an episodic expert demonstration $\xi_{\pi^{demo}}^k$, whose feature value is expected to be the same as the queried feature point $\varphi_{query}$. Both the learning policy roll-out $\xi_{\pi}^i$ and the expert episodic demonstration $\xi_{\pi_{demo}}^k$ will be added to the reply buffer $D$ to update agent policy using SAC as the underlying RL algorithm. We summarize the pseudo-code in Algorithm \ref{algorithm1}.

\begin{algorithm}
\caption{Episodic Active Learning from demonstration querY (EARLY)}\label{alg:cap}
% \hspace*{\algorithmicindent} \textbf{Input:} max length of explore history $N_h$, ratio threshold $r_{query}$, uncertainty measurement function $M(\cdot)$, feature function $\Phi (\cdot)$
\begin{algorithmic}[1]
\Require training iteration budget $i_{max}$, demonstration query budget $N_d$, max length of recent explored feature history $N_h$, ratio threshold $r_{query}$, uncertainty measurement function $M(\cdot)$, feature function $\Phi (\cdot)$

\State Initialize Q-value nets $Q_{\theta_k, k \in \{1, 2\}}$, value net $V_{\psi}$, target value net $V_{\bar{\psi}}$, policy net $\pi_{\phi}$
\State Initialize replay buffer $D$
\State Initialize feature history $H$, feature uncertainty history $H_u$
\State $idx_{thres} \gets N_h \times r_{query}$
\State $queried \; demo \gets 0 $
\For{iteration $i \in \{1, 2, ...\}$}
    \State rollout the policy $\pi$ to get an episodic trajectory $\xi_{\pi}^i$
    \State calculate the corresponding feature value $\varphi_i = \Phi(\xi_{\pi}^i)$
    % \State uniformly sample a feature point $\varphi_i \in \Phi$
    % \State recover learner's roll-out trajectory $\xi_i \gets \Phi^{-1}_{\pi}(\varphi_i | \pi_{\phi})$
    \For{step $t \in \xi_{\pi}^i$}
        \State update $D$, $Q_{\theta_k}$, $V_{\psi}$, $V_{\bar{\psi}}$, $\pi_{\phi}$
    \EndFor
    \State calculate feature uncertainty $u_i \gets M(\xi_{\pi}^i, Q_{\theta_k}, V_{\psi}, V_{\bar{\psi}}, \pi_{\phi})$
    \State update $H$ and $H_u$
    \If{size of $H >= N_h + 1$}
        \State ordered uncertainty history $H_u^{dsc} \gets Desc Order (H_u)$  
        \State $u_{thres} \gets H_u^{dsc}[idx_{thres}]$
         \If{$u_i > u_{thres}$ and $queried \; demo < N_d$}
             \State feature to query $\varphi_{query} \gets \argmax_{\varphi_j \in H}H_u$
             \State get an expert demo $\xi_{\pi^{demo}}^k$ of feature value $\varphi_{query}$
             \State update $D$, $Q_{\theta_k}$, $V_{\psi}$, $V_{\bar{\psi}}$, $\pi_{\phi}$
             \State $queried \; demo \gets queried \; demo + 1$
        \EndIf
        \State remove the earliest element from $H$ and $H_u$
    \EndIf
\EndFor

\end{algorithmic}
\label{algorithm1}
\end{algorithm}

%%%%%%%%%%%%%%%%%%%%%%%%%%%%%%%%%%%%%%%%%%%%%%%%%%%%%%%%%%%%%%%%%%%%%%%%

\section{Experimental Setup}
To validate the effectiveness of our method, we tested on three simulated navigation tasks with sparse rewards, continuous state-action space, and increasing difficulty. We chose them as the testbed tasks since they are typical cases where a human demonstrator intuitively tends to know how to execute the task itself, but may not be optimal in teaching the task. Furthermore, their intrinsic long-horizon and spare-reward characteristics also make conventional RL algorithms more susceptible to converging to local optimum, making these tasks a more challenging scenario to test algorithm performance. We first conducted experiments with simulated oracle demonstrators to evaluate the learning performance of our method against other baselines. Furthermore, we also conducted a pilot user study with human expert demonstrators ($N=18$) to prove the learning efficacy of our method for real human users and investigate its user experience in terms of perceived task load and human time cost.

% We chose to test our method on tasks of sparse rewards since this type of tasks is more common in real-world applications, where the policy performance is dominantly dependent on whether a goal is achieved or not. At the same time, their intrinsic long-horizon and spare-reward characteristics also make conventional RL algorithms more susceptible to converging to local optimum, making these tasks a more challenging scenario to test algorithm performance. We compared our methods with several other baselines in policy learning performance, aiming to discover the potential benefits our method may bring to agent policy learning.

\subsection{Task Environments}
\begin{figure}[t!]
  \centering
  \begin{subfigure}[b]{0.32\linewidth}
    \includegraphics[width=\linewidth]{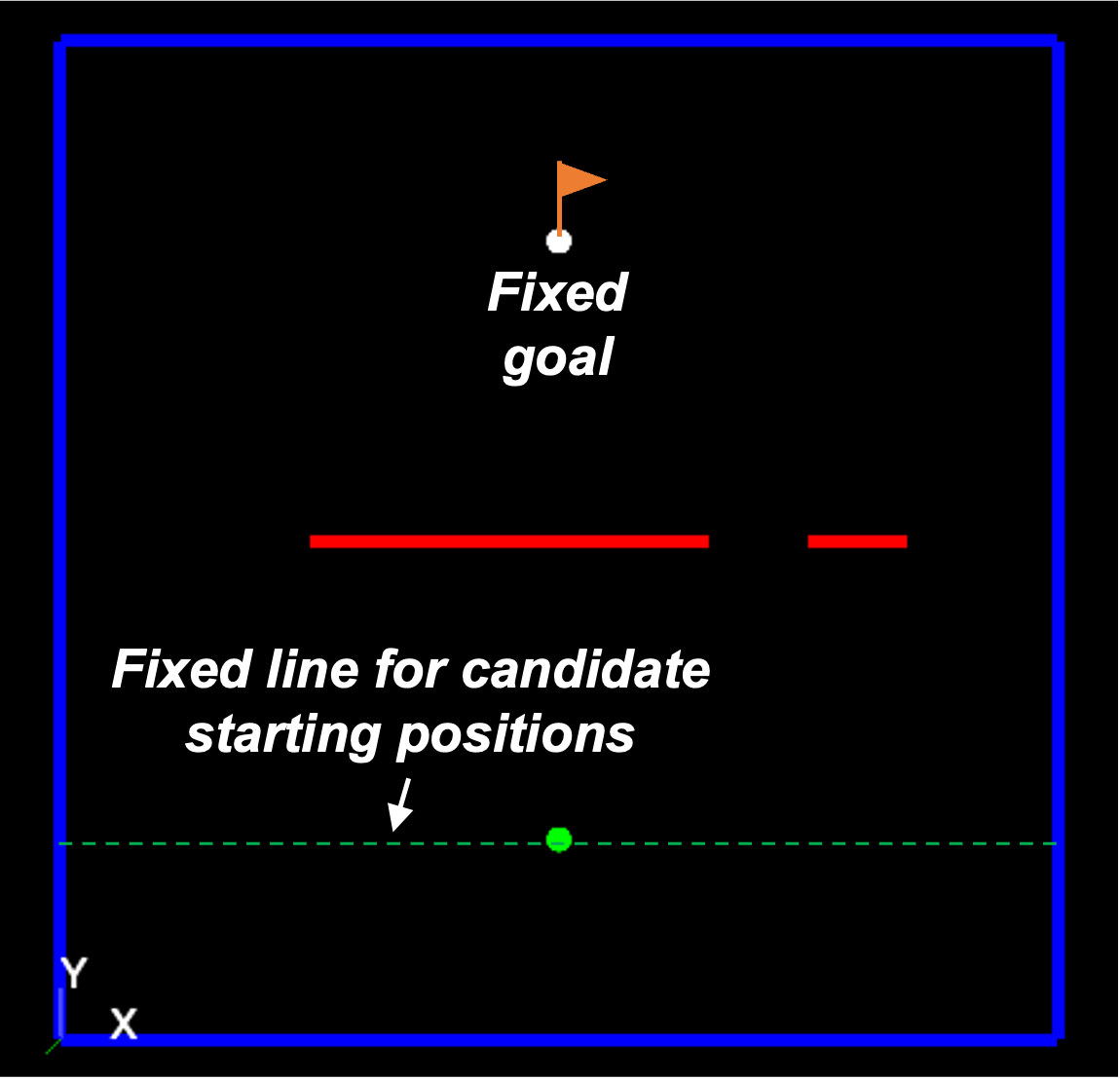}
     \caption{nav-1}
  \end{subfigure}
  \begin{subfigure}[b]{0.32\linewidth}
    \includegraphics[width=\linewidth]{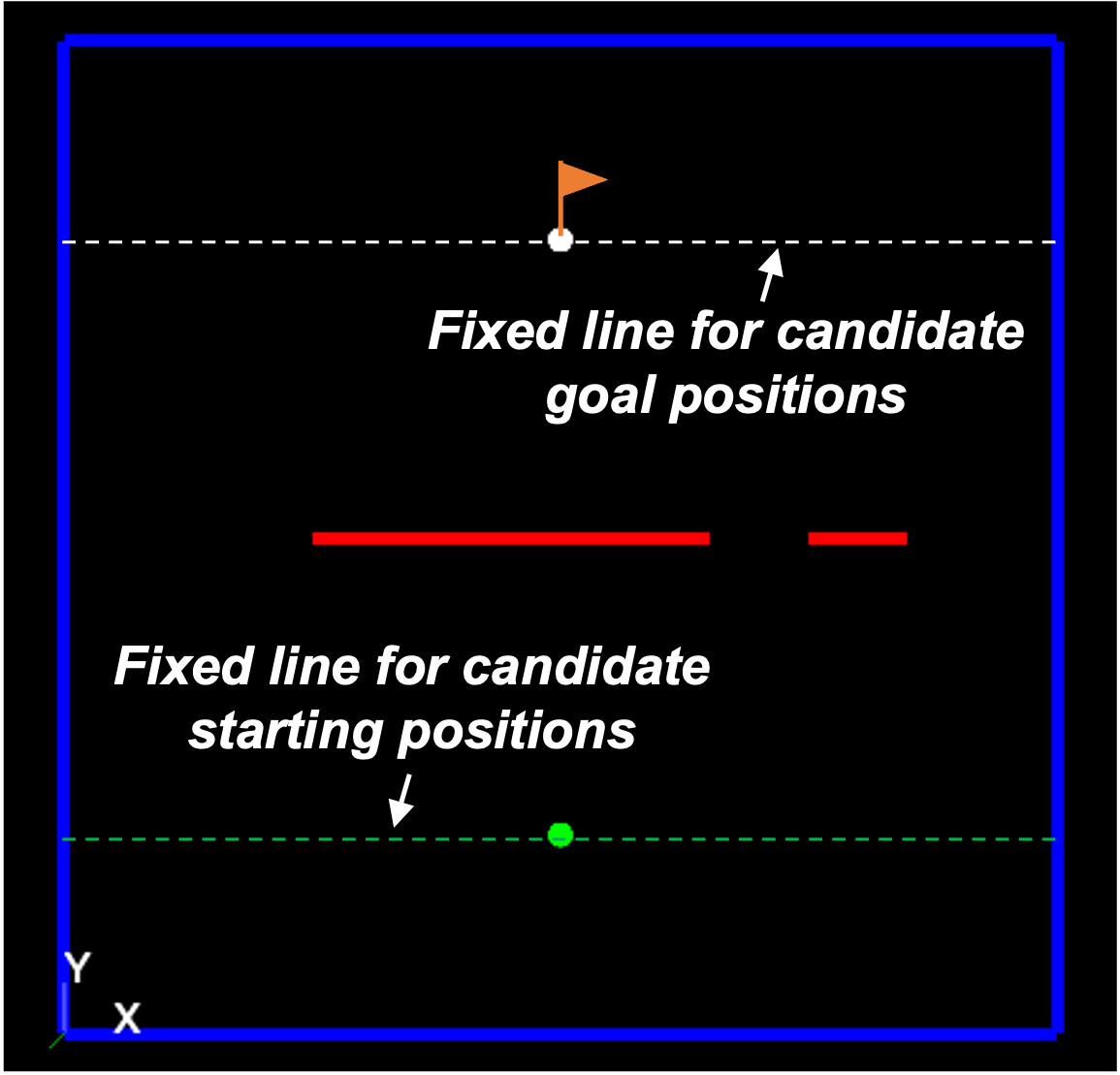}
    \caption{nav-2}
  \end{subfigure}
  \begin{subfigure}[b]{0.32\linewidth}
    \includegraphics[width=\linewidth]{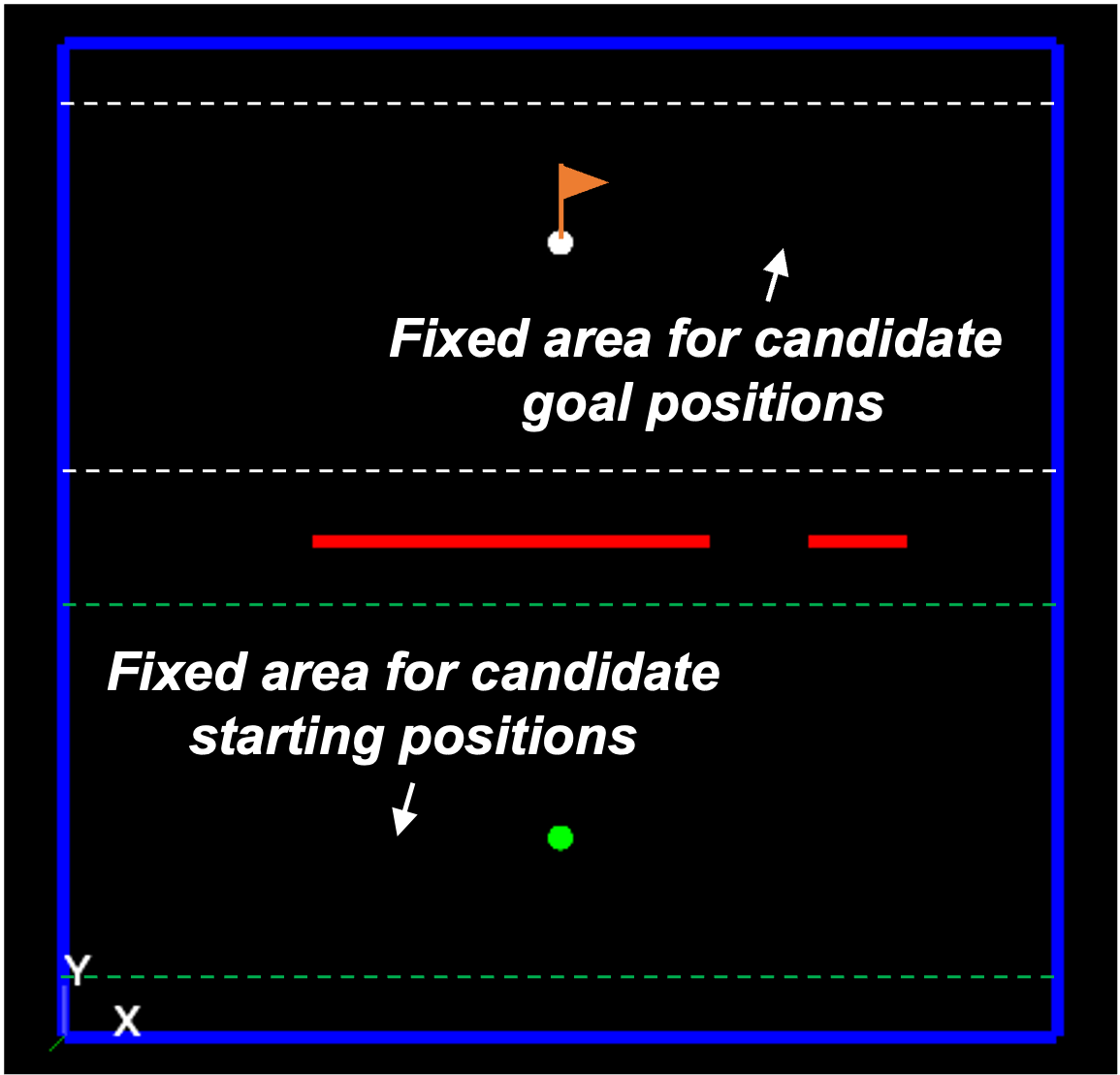}
    \caption{nav-3}
  \end{subfigure}
  
  \caption{Task environments for three simulated navigation tasks of scaling difficulties.}
  \label{nav environments}
\end{figure}

We designed three simulated navigation tasks shown in Figure ~\ref{nav environments}. For each task, we defined the state $s_t$ as $s_t=(x_t, y_t, x_{goal}, y_{goal})$, where $(x_t, y_t)$ is the current position of the moving agent and $(x_{goal}, y_{goal})$ is the position of the destination. We defined the action $a_t$ as $a_t = (v_x, v_y)$, where $v_x, v_y \in [-1.0, 1.0]$ represent the agent moving velocity along the $x$ and $y$ axis at step $t$. The agent will receive a reward $r_t$ of $-1$ after each step, a reward of $-1000$ if it bumps into the map boundary or obstacles, or a reward of $1000$ if it arrives near the goal within a distance of $1.0$ unit. An episode will terminate once the agent bumps into map boundary or obstacles, arrives at the destination area, or it reaches the maximum episode length of $200$ steps. 

More specifically, these three navigation tasks are of increasing difficulty. For the task of 
fixed-goal navigation (i.e., \textit{nav-1}), the agent aims to arrive at a fixed goal position with its initial positions randomly chosen from a fixed horizontal line. For the task of random-goal navigation (i.e., \textit{nav-2}), both the initial positions and the goal positions will be randomly chosen from a horizontal line before each episode starts. For the task of advanced random-goal navigation (i.e., \textit{nav-3}), the initial positions and the goal positions will be randomly chosen from two areas, leading to an increasingly larger search space for policy learning from nav-1 to nav-3.

% All three navigation tasks share a similar objective: the agent is required to navigate from certain points located in the lower half of the map, avoid obstacles, and arrive at some destination located in the upper half of the map as soon as possible. However, the distributions of the initial agent positions and destination positions are different among the three tasks. For the task of \textit{fixed-goal navigation}, the agent aims to arrive at a pre-defined fixed destination. This destination position will remain the same for all episodes. By contrast, the initial position of the agent will be randomly assigned along a fixed horizontal line (i.e., random $x_{0}$ and fixed $y_{0}$) before each episode starts. Similarly, in the task of \textit{random-goal navigation}, the initial position of the agent will still be randomly assigned along the same predefined fixed horizontal line. However, the destination position will also be randomly decided along another fixed horizontal line (i.e., random $x_{goal}$ and fixed $y_{goal}$). And this will lead to a much larger search space for the task and pose more challenges for policy learning. For the task of \textit{advanced random-goal navigation}, the initial position and the destination position will still be randomly assigned before each episode starts, but each of them will be sampled from an area as opposed to a horizontal line (i.e., random $(x_{0}, y_{0})$ and random $(x_{goal}, y_{goal})$), which makes the search space of the task even larger. 

\subsection{Baselines}
To evaluate how our method may benefit agent policy learning, we compared our method with $4$ other baselines:
\begin{enumerate}
    \item \textbf{DDPG-LfD}: a popular method for reinforcement learning from demonstrations \cite{vecerik2017leveraging}. The agent learns in a conventional ``demo-then-training'' manner, where episodic expert demonstrations are first randomly collected and added to the reply buffer before the learning agent starts to update its control policy using DDPG.
    \item \textbf{I-ARLD}: a state-of-the-art method that learns in a ``demo-while-training'' manner \cite{chenactive}. It switches control from the learning agent to the expert demonstrator during the agent roll-outs, resets the environment to previous moments, and only queries \textit{isolated} state-action pairs for the next few steps before switching control back to the learning agent.
    \item \textbf{GAIL}: a classic imitation learning algorithm that also learns in a ``demo-then-training'' manner \cite{ho2016generative}.
    \item \textbf{BC}: one of the most common imitation learning algorithms that directly treats policy training as a conventional supervised learning problem \cite{pomerleau1988alvinn}.
\end{enumerate}

\begin{figure*}[t!]
  \centering
  
  \begin{subfigure}[b]{0.33\linewidth}
    \includegraphics[width=\linewidth]{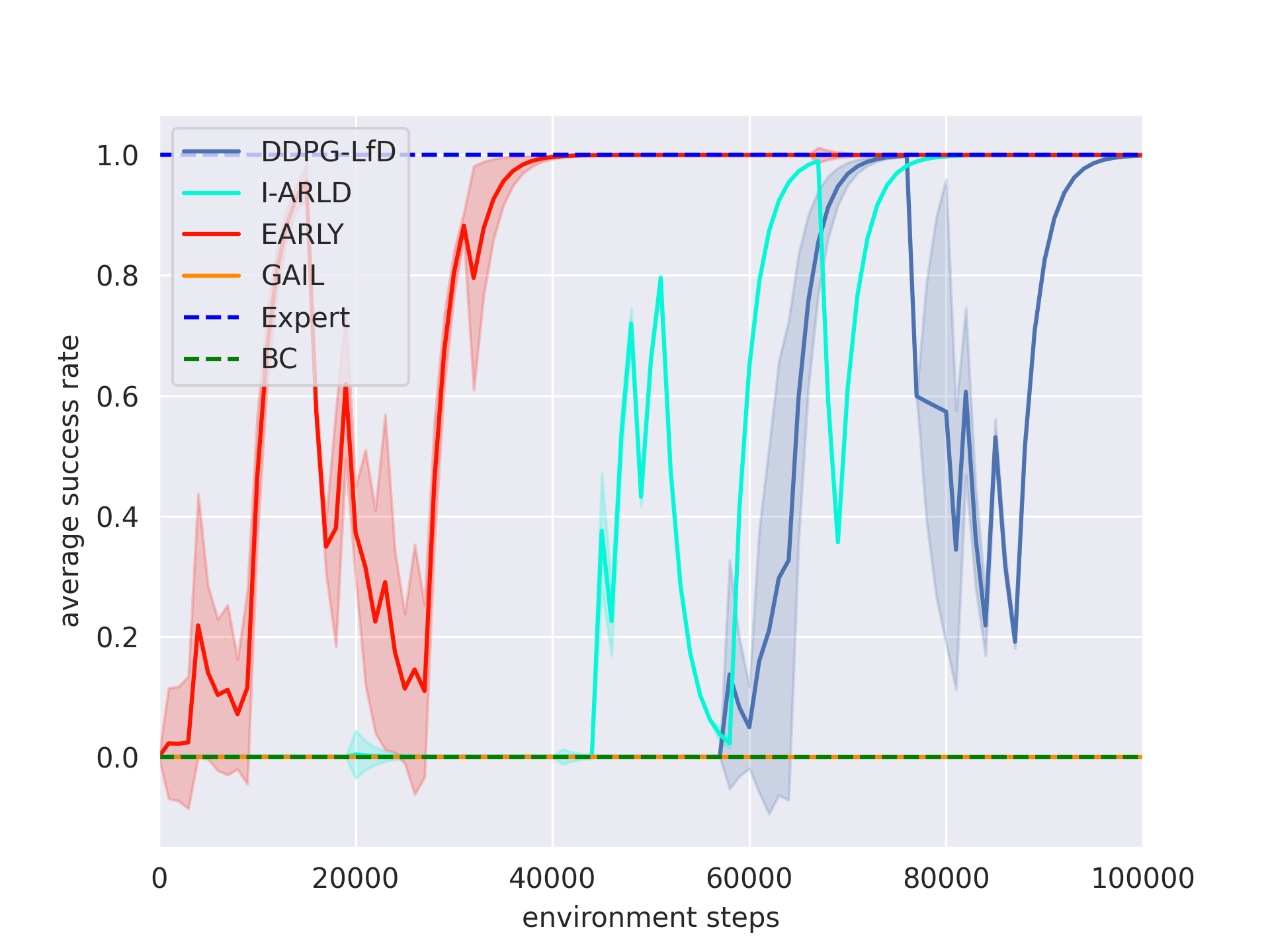}
     \caption{nav-1}
  \end{subfigure}
  \begin{subfigure}[b]{0.33\linewidth}
    \includegraphics[width=\linewidth]{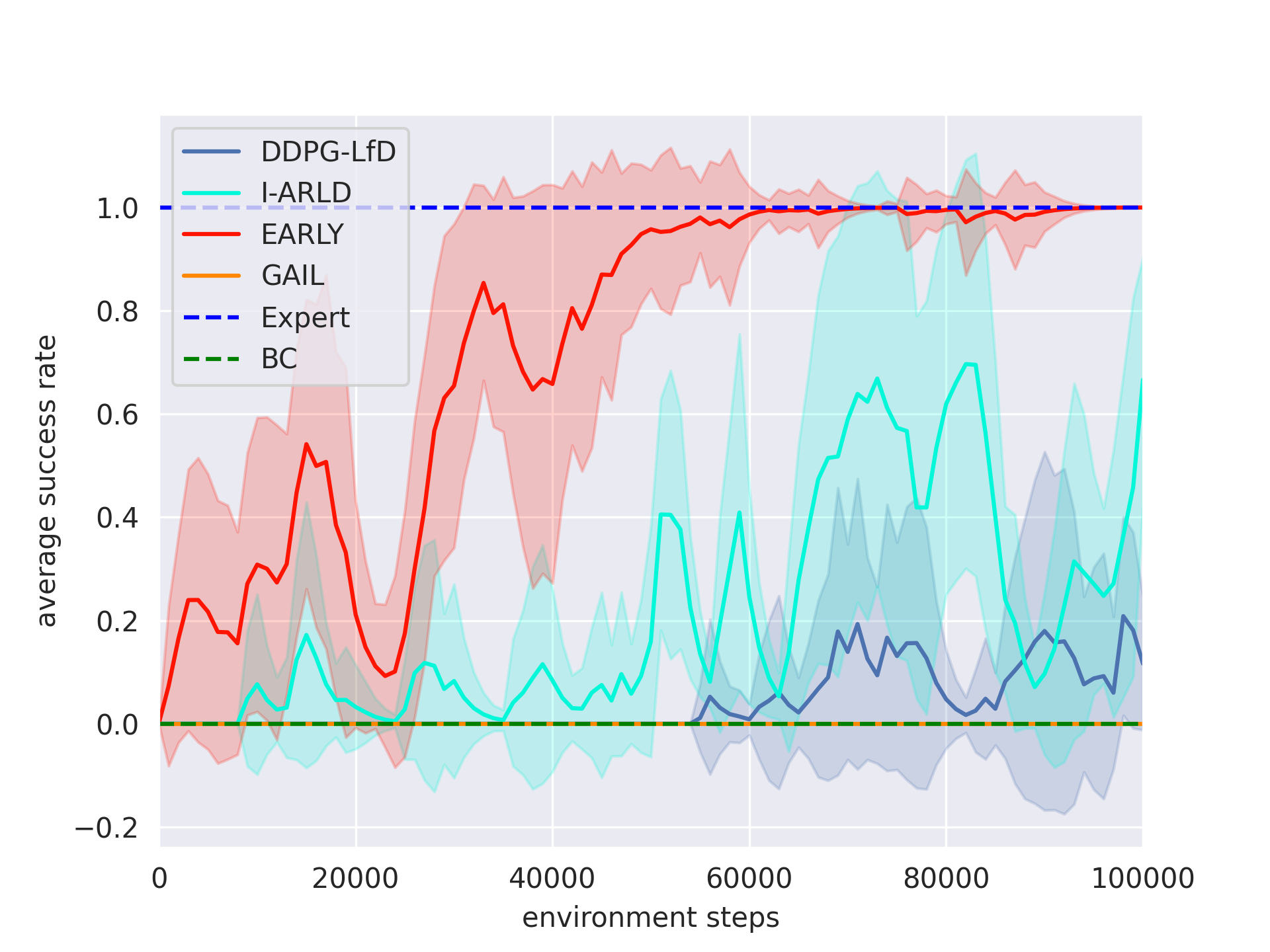}
    \caption{nav-2}
  \end{subfigure}
  \begin{subfigure}[b]{0.33\linewidth}
    \includegraphics[width=\linewidth]{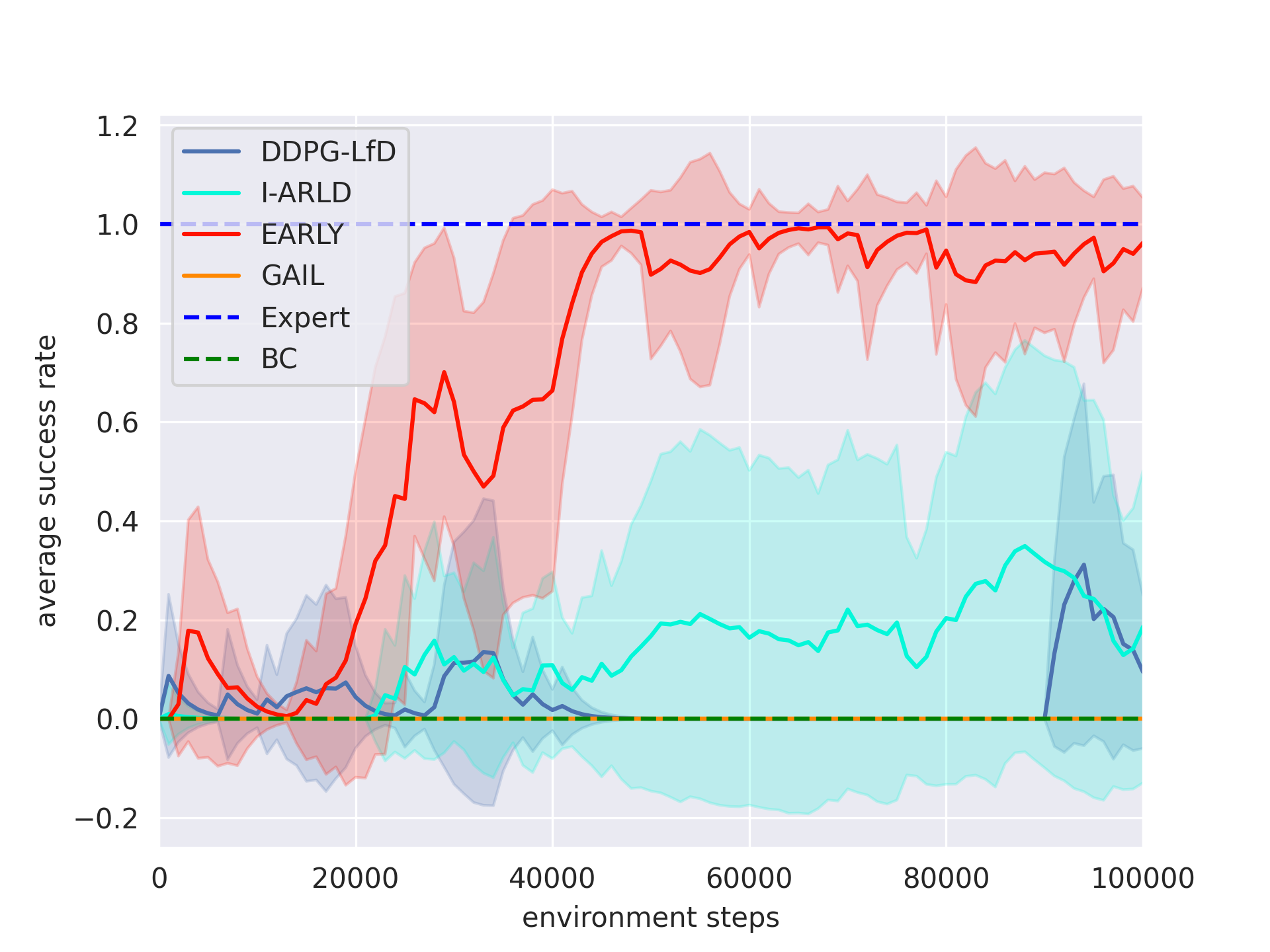}
    \caption{nav-3}
  \end{subfigure}
  
  \caption{Results of the experiments with simulated-oracle demonstrators. The shaded areas represent the standard deviation.}
  \label{reward and success plot for nav}
\end{figure*}

% \begin{enumerate}
%     \item \textbf{B-ARLD}: it decides to query for episodic expert demonstrations at a fixed probability $p_b$ that follows a Bernoulli distribution, but to query the most uncertain feature point in the shifting recent history.
%     \item \textbf{R-ARLD}: it follows the same strategy as E-ARLD to decide when to query, but randomly samples a feature point from the feature space to query for demonstrations.
%     \item \textbf{P-RLD}: the learning agent passively uses reinforcement learning from demonstrations. All episodic expert demonstrations are first randomly collected and added to the reply buffer before the learning agent starts to update its control policy using SAC.
%     \item \textbf{PH-RLD}: the learning agent follows the same procedures as P-RLD, except that episodic expert demonstrations are from real human experts.
%     % \item \textbf{GAIL}: the agent updates its control policy using Generative Adversarial Imitation Learning (GAIL) ~\cite{ho2016generative} without receiving the reward signal from the environment. All expert demonstrations are randomly collected before the agent starts to update its policy.
%     \item \textbf{SAC}: the agent updates its control policy simply using SAC without any expert demonstrations.
% \end{enumerate}

For our method, we chose the ratio threshold $r_{query}$ as $0.35, 0.4$, and $0.55$ for three navigation tasks respectively, and set the maximum length of recent explored history $N_h$ as $20$. For the underlying SAC algorithm, we followed the same settings of neural network structures, hyperparameters, and the optimizer as in ~\cite{haarnoja2018soft}. For DDPG-LfD and I-ARLD, we reproduced them according to their original papers with the default parameters. For GAIL and BC, we implemented them using the open-source library \cite{gleave2022imitation} for stable implementation. For all baselines, we trained the policy with $1 \times 10^5$ environment steps for all three tasks respectively.

Additionally, we did not find performance improvement by using Prioritized Experience Replay (PER) ~\cite{schaul2015prioritized} for the reply buffer. Instead, we maintained two separate reply buffers for current policy roll-outs and expert demonstrations. To guarantee the expert demonstrations can be stably sampled, we sampled the same amount of roll-outs from expert demonstrations as those from the agent policy to comprise each sampling batch. All expert demonstrations will be stored in the corresponding reply buffer through the whole learning process, while the earliest agent roll-out will be removed from the reply buffer for the agent policy once it exceeds the buffer capacity.

\subsection{Experiments with Oracle-Simulated Demonstrators}
We first conducted experiments using oracle-simulated demonstrators to evaluate the learning performance of our method. We used RRT* ~\cite{karaman2010incremental}, a state-of-the-art path planning algorithm, as the oracle to provide episodic demonstrations upon receiving feature-oriented queries from the learning agent. Since we chose the \textit{initial state} as the feature $\varphi_i$ of a given episodic roll-out trajectory $\xi_{\pi}^i$,  whenever a feature query $\varphi_{query}$ (i.e., $s_0^{query}$) was generated, we intuitively used the RRT* algorithm to obtain an episodic expert roll-out trajectory that starts from $s_0^{query}$ and arrives at the destination. For the baselines that learn in a ``demo-then-training'' manner (i.e., DDPG-LfD, GAIL, and BC), we uniformly sampled from the initial state space to select the initial states of the expert demonstrations. To keep data collection labor aligned with a reasonable amount for real human demonstrators, we only allowed the learning agent to query $60$ episodic expert demonstrations (i.e., $N_d=60$) for each baseline method (or of an equal amount of total steps for I-ARLD).

% \subsection{Expert Demonstrations Collection}
% Aiming to investigate the learning performance and user experience of our method against other baselines, we designed two sections of experiments, one with an oracle-simulated expert and another one with real human experts. 

\begin{figure*}[t!]
  \centering
  \begin{subfigure}[b]{0.22\linewidth}
    \includegraphics[width=\linewidth]{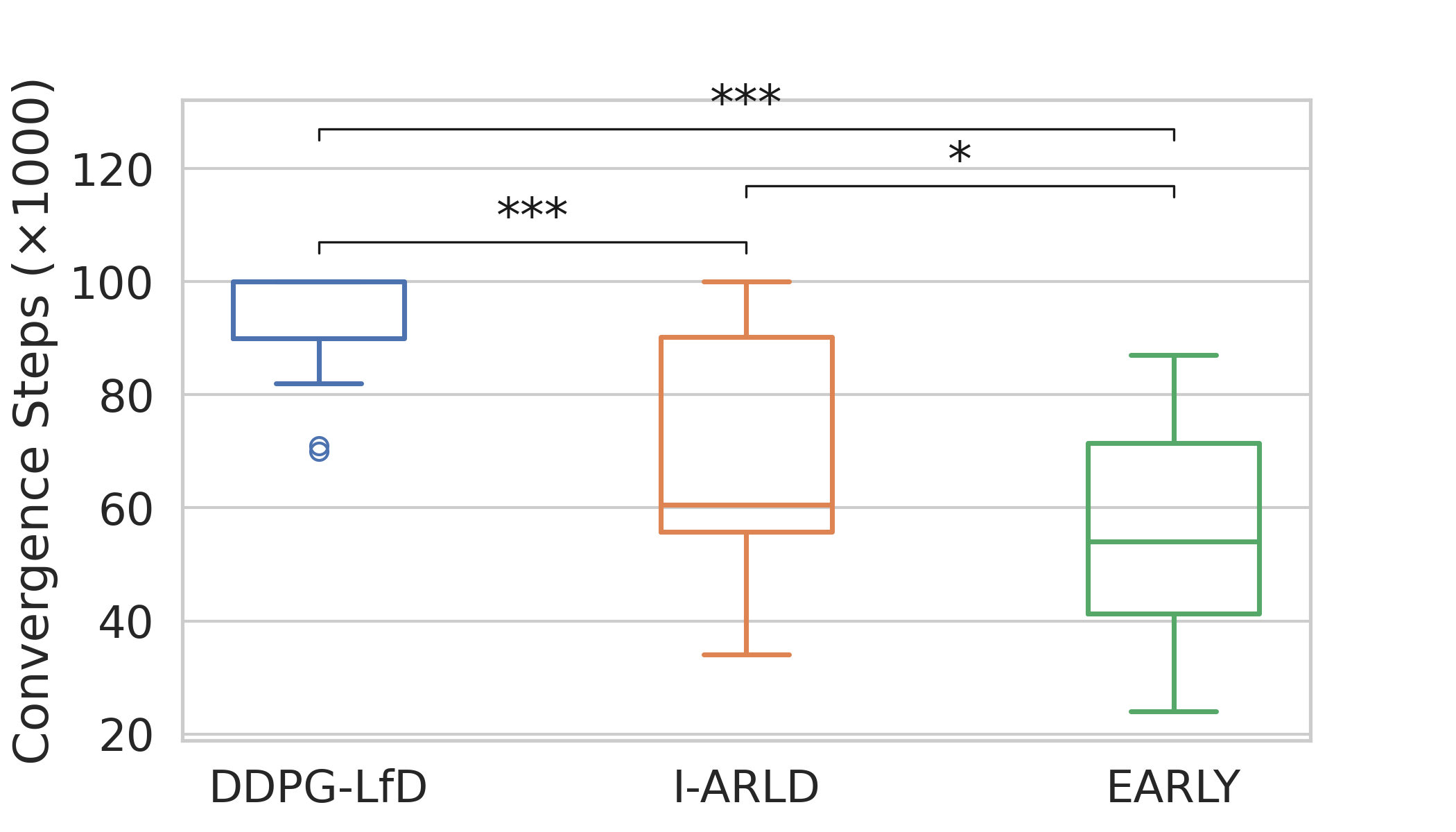}
    \caption{Steps to converge}
  \end{subfigure}
  \begin{subfigure}[b]{0.22\linewidth}
    \includegraphics[width=\linewidth]{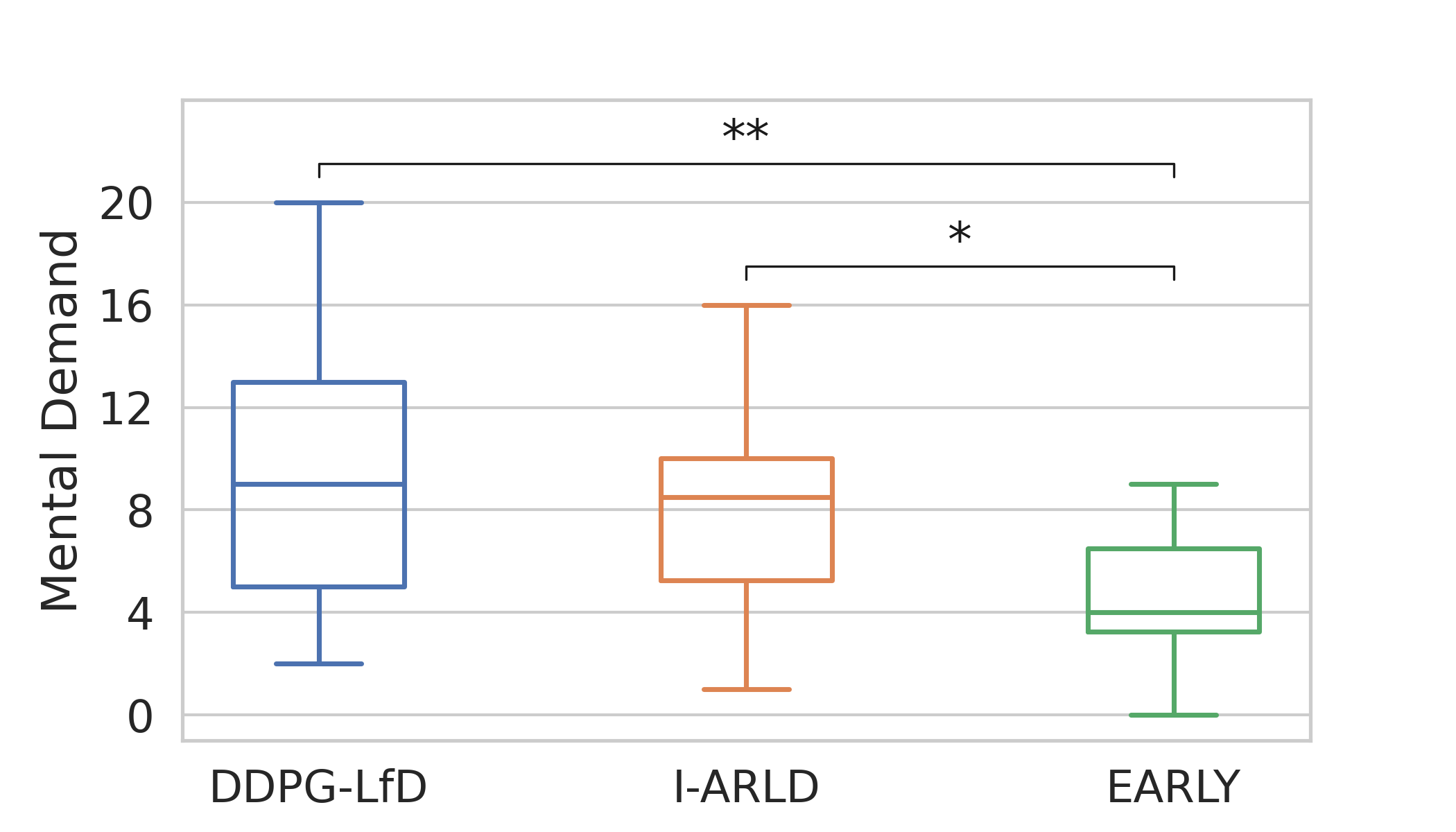}
    \caption{Mental demand}
  \end{subfigure}
  \begin{subfigure}[b]{0.22\linewidth}
    \includegraphics[width=\linewidth]{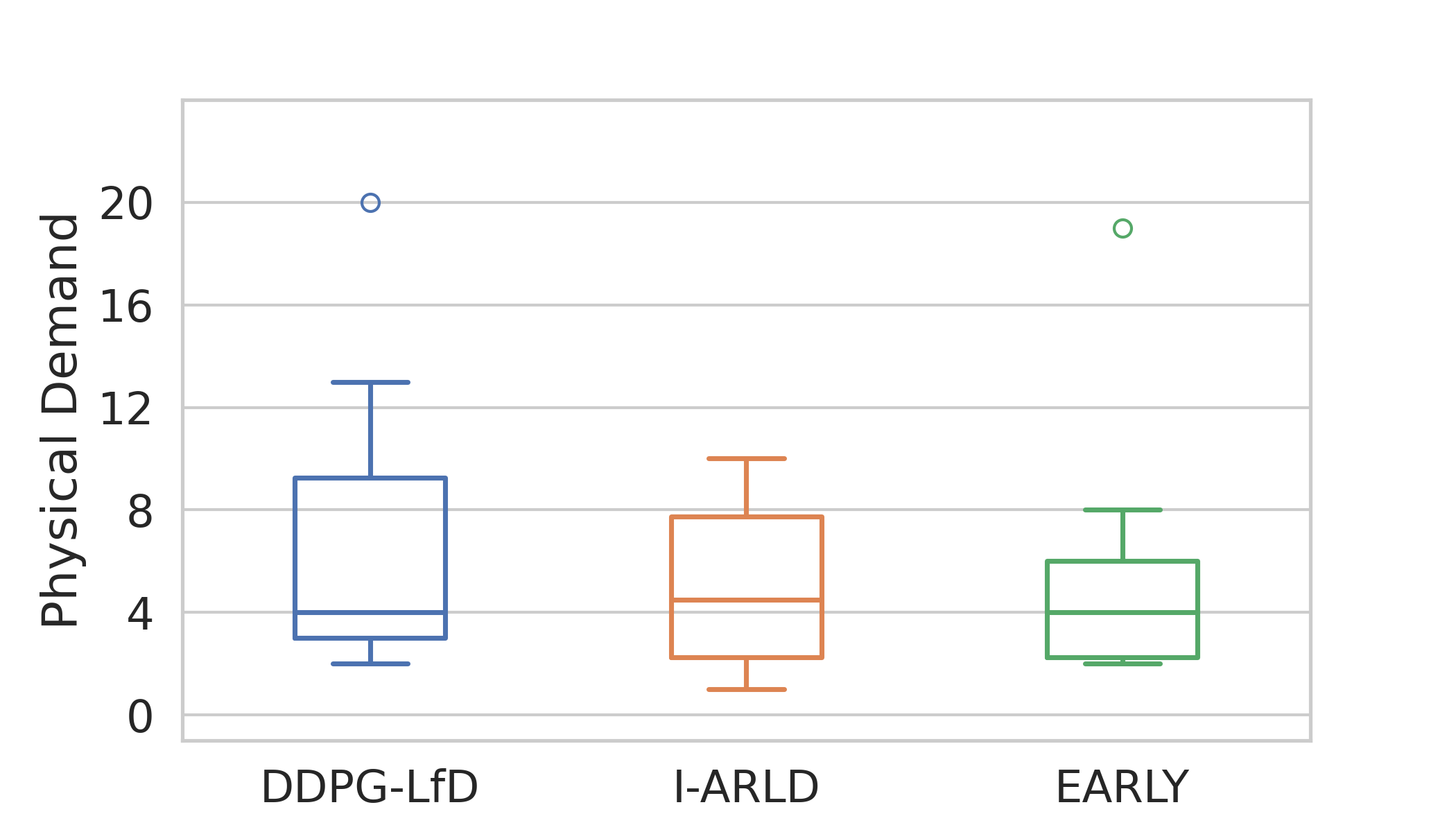}
    \caption{Physical demand}
  \end{subfigure}
  \begin{subfigure}[b]{0.22\linewidth}
    \includegraphics[width=\linewidth]{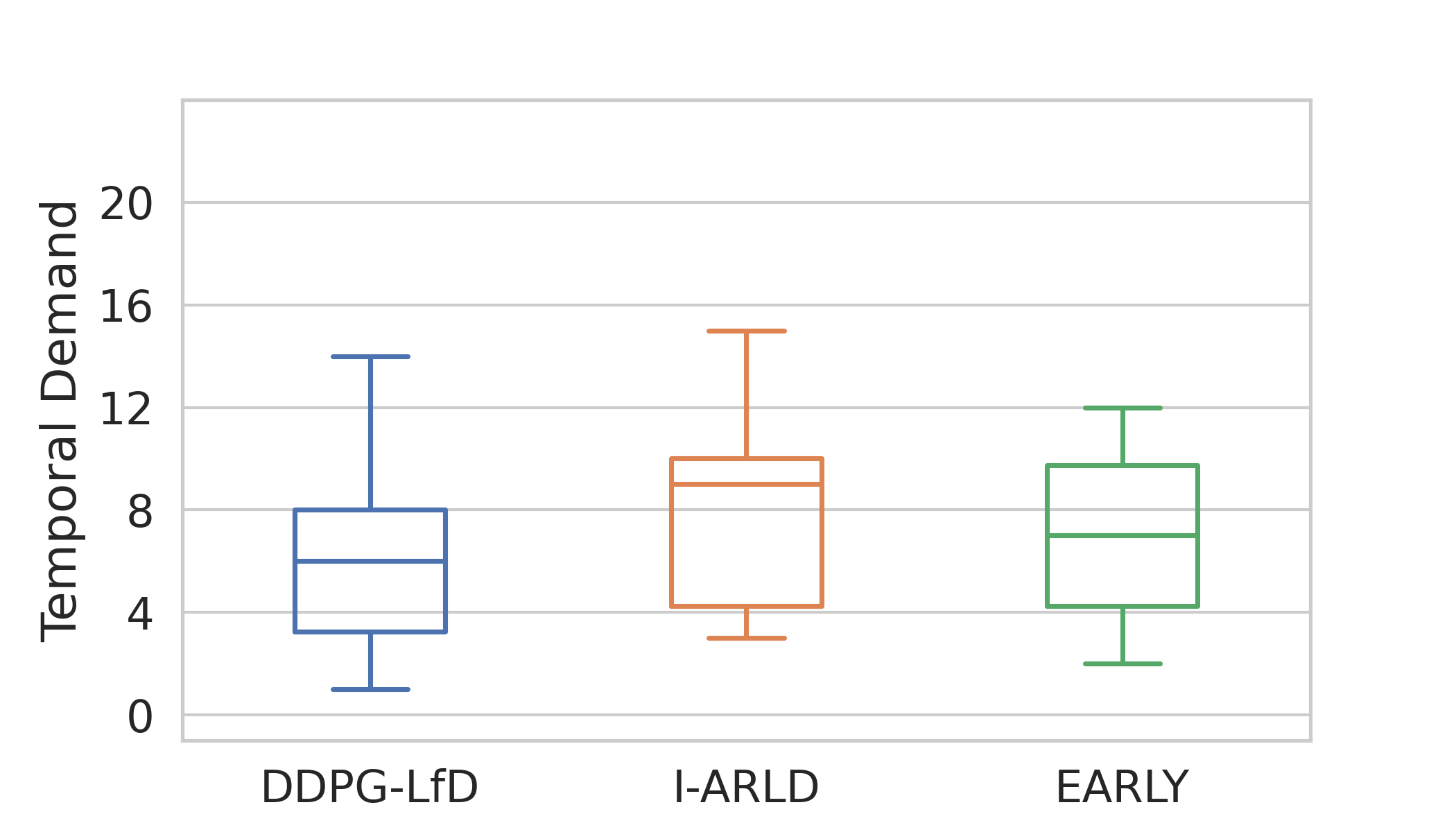}
    \caption{Temporal demand}
  \end{subfigure}
  \begin{subfigure}[b]{0.22\linewidth}
    \includegraphics[width=\linewidth]{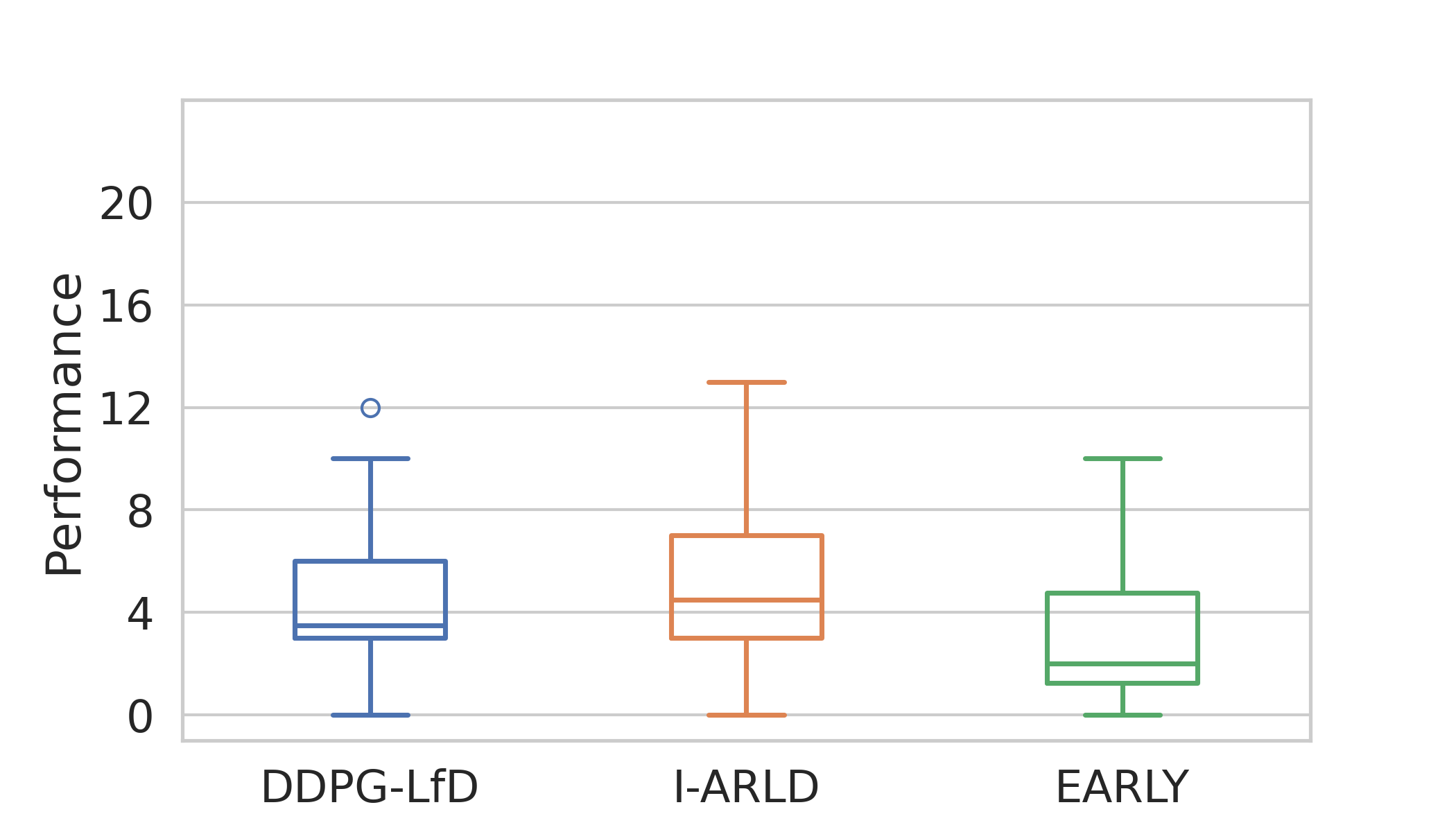}
    \caption{Perceived performance}
  \end{subfigure}
  \begin{subfigure}[b]{0.22\linewidth}
    \includegraphics[width=\linewidth]{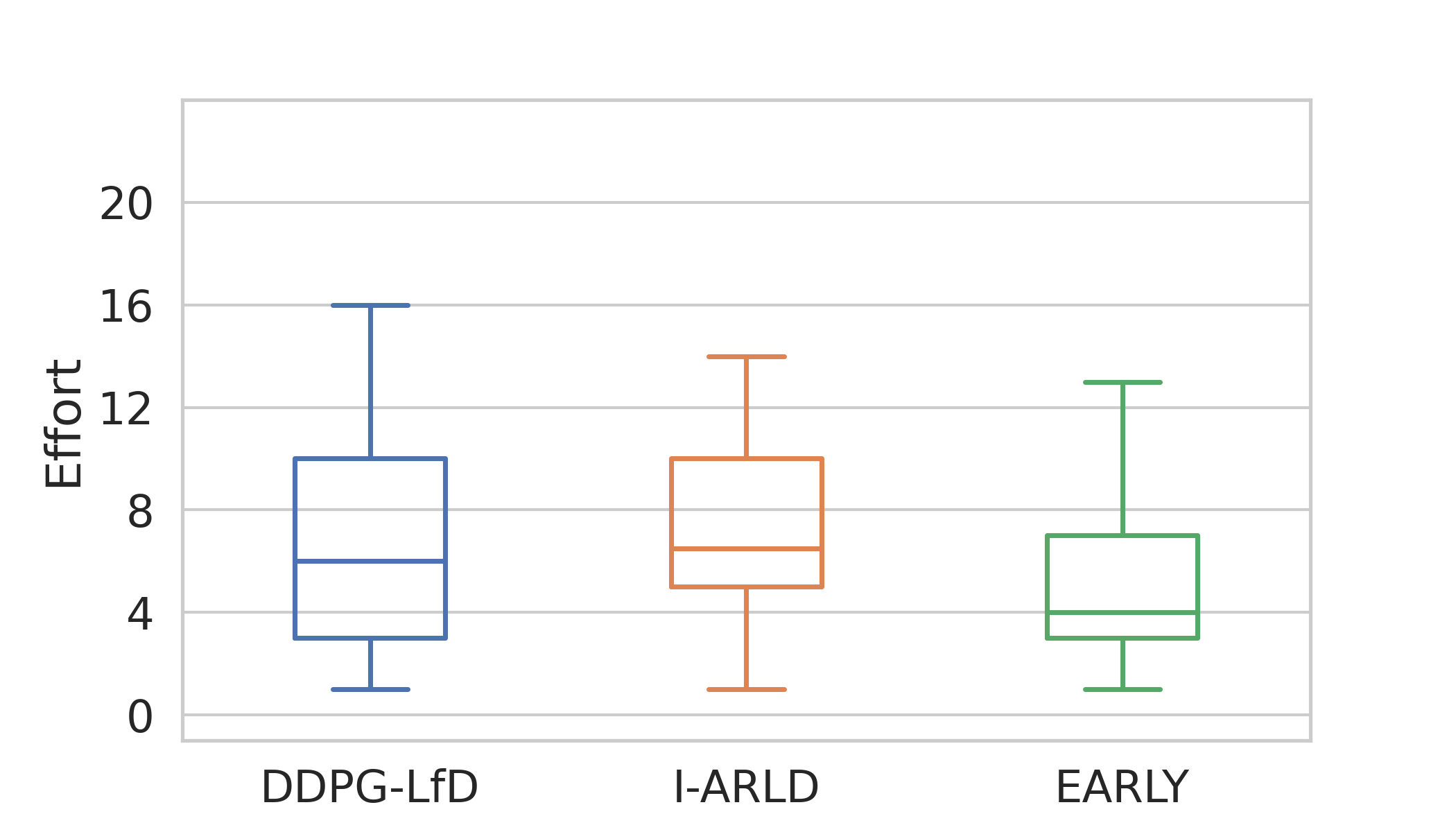}
    \caption{Effort}
  \end{subfigure}
  \begin{subfigure}[b]{0.22\linewidth}
    \includegraphics[width=\linewidth]{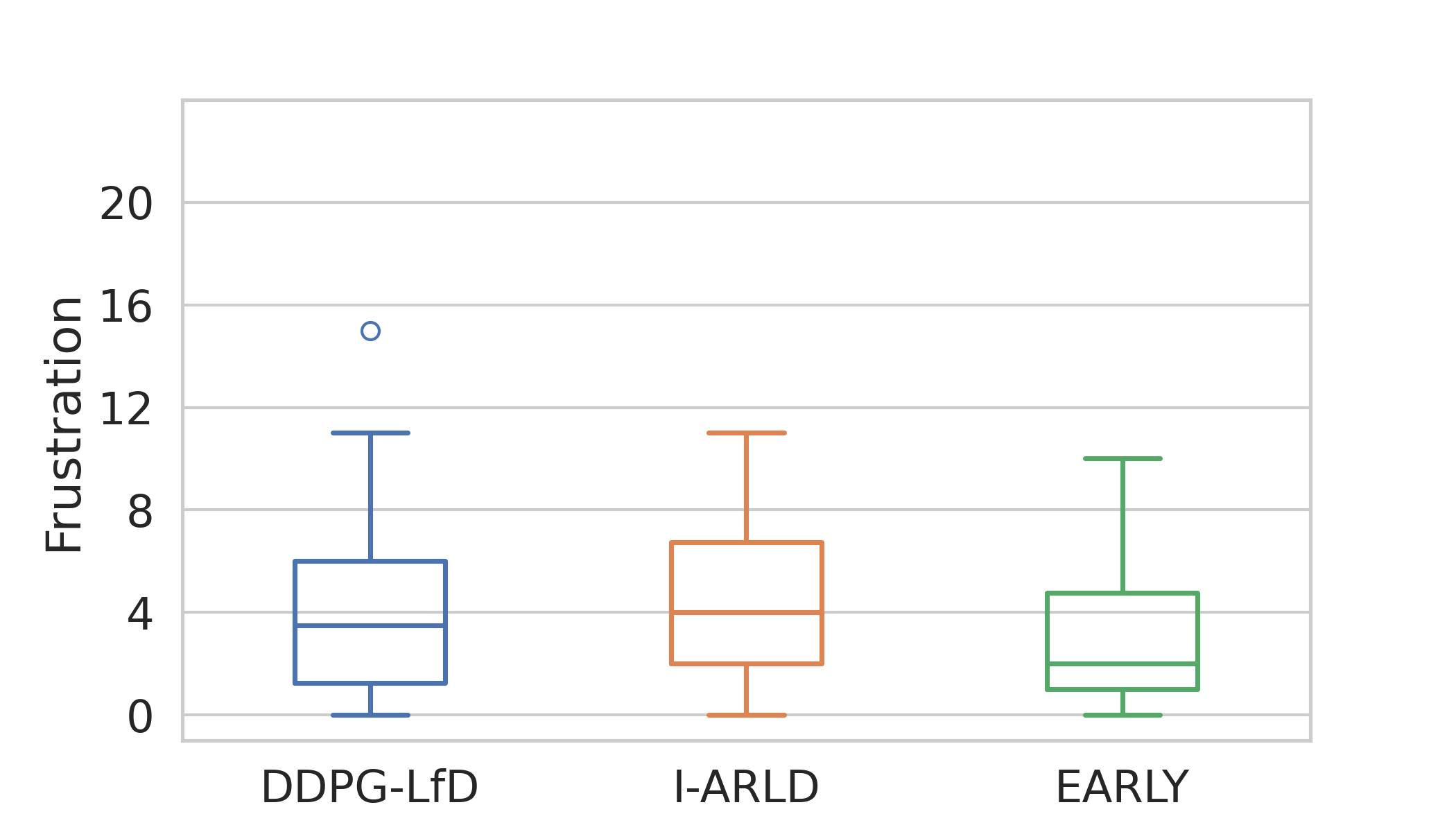}
    \caption{Frustration}
  \end{subfigure}
  \begin{subfigure}[b]{0.22\linewidth}
    \includegraphics[width=\linewidth]{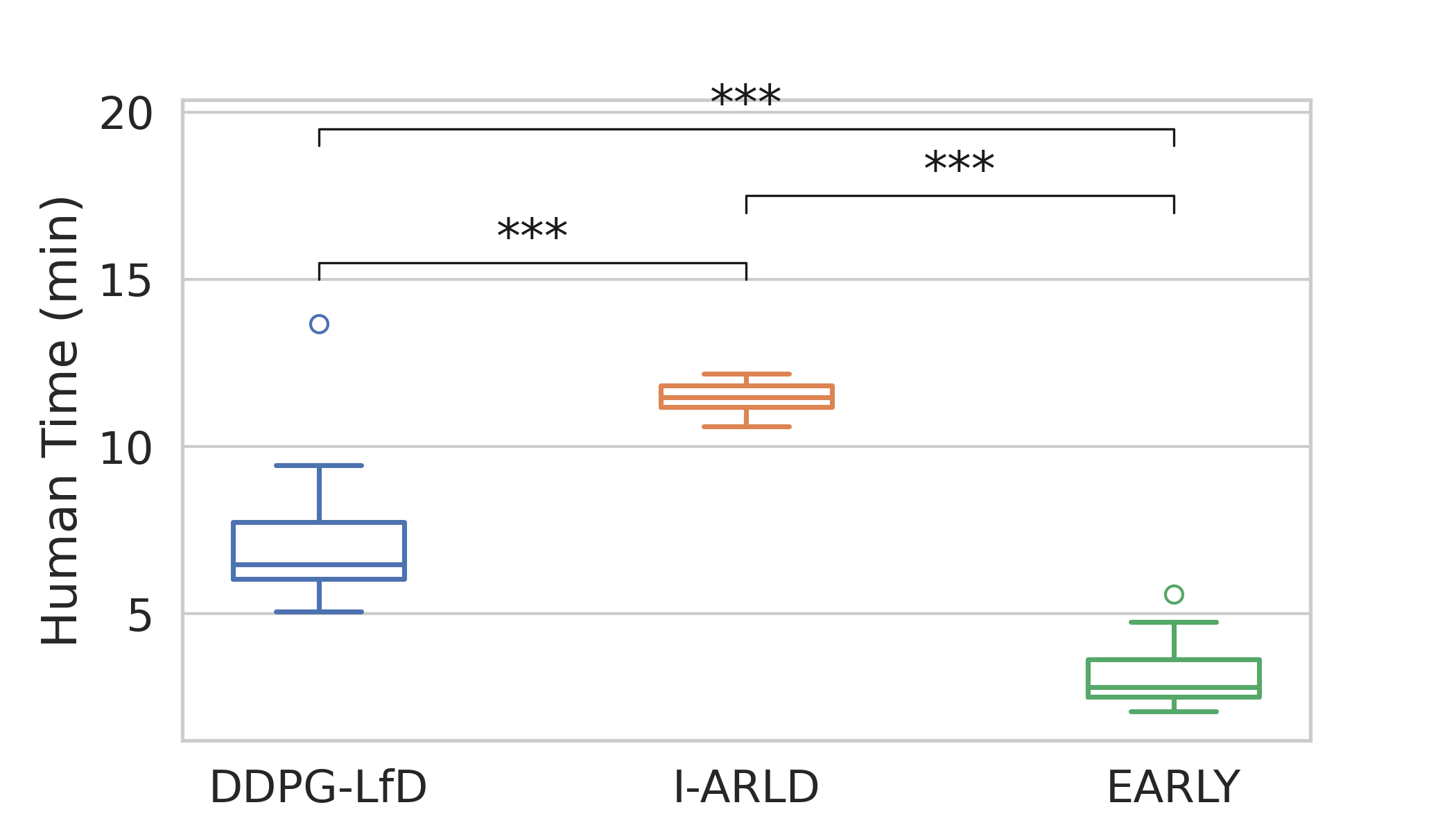}
    \caption{Human time}
  \end{subfigure}
  
  \caption{Results of experiments with real human demonstrators.}
  \label{user experience}
\end{figure*}

\subsection{Pilot User Study with Human Expert Demonstrators}
To investigate the efficacy of our algorithm and its user experience for real human users, we conducted a pilot user study with $18$ human participants ($9$ male, $8$ female, and $1$ other; $12$ aged between $18-29$ and $6$ aged between $30-39$; $11$ of some experience of machine learning and $7$ of extensive experience). We recruited them from campus via poster advertisement with the approval of our faculty research ethics board. We obtained their consent for experiments and data collection before the experiments began and compensated for their participation with a \euro{$10$} gift card.

Participants will go through $3$ different methods for demonstration collection (i.e., DDPG-LfD, I-ARLD, and EARLY) for the task of nav-1 in a counter-balanced order. Each participant will use a joystick to provide $60$ episodic demonstrations (or of an equal amount of total steps for I-ARLD) using each of these methods. For the method of DDPG-LfD, we conducted demonstration collection as an unguided demo-then-training process. Participants will follow their own strategies to choose the starting positions of their demonstrations that they believe to be most beneficial for agent learning, and use the joystick to provide complete demonstrations to navigate from their chosen starting positions to the fixed goal position. For the other two methods, we conducted data collection as a guided demo-while-training process. The learning agent will utilize its own query strategy to determine the position it needs help with, and participants will then use the joystick to navigate it from the queried position to the fixed goal position. 

To evaluate the user experience of each method, participants will fill out a standard NASA-TLX questionnaire to quantify their perceived workload after the experiment section of each method. For each participant, we also counted the total amount of human time spent for each method, starting from the experiment began until all $60$ demonstrations were provided. Furthermore, we designed an open-ended question after the experiments of DDPG-LfD to ask about each participant's strategy when choosing their demonstrations. Before all the experiments started, there was a training session of up to 5 minutes. It finished after the participant succeeds in navigating the agent to the goal position $5$ times in a row, or it reaches the 5-minute limit.

% By contrast, to collect expert demonstrations for all navigation tasks with real human experts, we invited $5$ human experts for experiments. However, considering that this work is mainly focused on the \textit{sub-optimality in teaching the task} as opposed to the \textit{sub-optimality in executing the task}, we only let the human experts choose the feature values $\varphi$ (i.e., initial states $s_0$) of the episodic expert demonstrations they would like to provide. After they finished choosing the initial states of all the expected demonstrations, we then used the RRT* to plan the paths that started from these human-chosen initial states and arrived at the destination. These planned trajectories were collected as real human expert demonstrations and only used to train the policy for the baseline PH-RLD. To facilitate potential usage for human experts with reasonable data collection labor, for both types of demonstration collection processes, we only allowed the learning agent to query $60$ episodic expert demonstrations (i.e., $N_d=60$) for each baseline method.

%%%%%%%%%%%%%%%%%%%%%%%%%%%%%%%%%%%%%%%%%%%%%%%%%%%%%%%%%%%%%%%%%%%%%%%%

\section{Results and Discussion}

\subsection{Experiments with Oracle Experts}
To evaluate the learning performance, we calculated the average success rate over $1000$ test episodes at an interval of $1000$ environment steps during the policy training process. The initial states of these test episodes were randomized using different random seeds. 

As shown in Figure ~\ref{reward and success plot for nav}, DDPG-LfD and I-ARLD only managed to converge to the expert-level performance for the task of nav-1 at around $9.7 \times 10^4$ and $8 \times 10^4$ environment steps. For the task of nav-2 and nav-3, both of them only reached sub-optimal performance that was much worse than the expert. By contrast, our method achieved expert-level performance for all three tasks. Furthermore, our method only took around $4 \times 10^4$ steps to converge to the expert-level performance in the task of nav-1, which is over $58.7\%$ and $50.0 \%$ faster than DDPG-LfD and I-ARLD respectively. For the method of GAIL and BC, neither of them managed to solve any of the navigation tasks within the given amount of environment steps.

As indicated by these results, what set of expert demonstrations to provide did have a large influence on the agent policy learning. The conventional paradigm of RLED where the learning agent passively receives and learns from the expert demonstrations may not best benefit policy learning. Moreover, when the demonstrator employs the uniform strategy of providing demonstrations, it may neglect how differently each area in the feature space contributes to the policy learning. By contrast, by actively evaluating agent uncertainty and querying for episodic target demonstrations, critical situations are more likely to encounter and acquire more attention from the demonstrator, leading to faster convergence to the expert-level performance.

\subsection{Experiments with Human Experts}

\subsubsection{Learning Performance}

Similarly, we trained navigation policies for each participant using the demonstrations collected by different baseline methods. During the training process, we measured the average success rate over $1000$ randomly initialized test episodes at an interval of $1000$ environment steps. We conducted a one-way repeated ANOVA test to investigate the effect of different learning algorithms on the convergence of success rate measured by environment steps. As shown in Figure \ref{user experience}, there was a significant difference in the convergence of success rate among different learning algorithms ($F(2, 34)=24.62, p<.001$) with a large effect size ($\eta^2=0.49$). The Tukey HSD post hoc test indicated that the success rate of EARLY ($M=53.94, SD=19.21$) converged significantly faster than DDPG-LfD ($M=93.28, SD=10.16$) and I-ARLD ($M=69.11, SD=20.14$). Furthermore, I-ARLD also shows a significantly faster convergence compared with DDPG-LfD. Complied with the results of experiments with simulated oracle experts, these results indicate that our method can still maintain efficacy when interacting with real human experts and benefit agent learning with faster convergence to the expert-level performance.

To further understand the reasons behind such a significant difference in learning performance, we looked into the participants' responses to the open-ended question that asked about their strategies in choosing what demonstrations to provide in the experiments of DDPG-LfD. $9$ of $18$ participants indicated that they tried to uniformly choose the starting positions, $2$ of them reported to have chosen the starting positions in a completely random manner, and $3$ of them indicated that they tried to uniformly choose the starting positions in the early phase and then shifted towards random ones. Additionally, $4$ of them reported that they were seeking to select ``critical'' starting positions that may have multiple equally optimal paths to the goal. As we can see from these results, even for such an intuitive navigation task, different human experts yet have quite diverse opinions on what distribution of demonstrations will most benefit agent learning. Such a discrepancy between how humans perceive the agent learning process and its actual learning process leads to wasting demonstrations of a limited budget on similar and redundant scenarios while neglecting more noteworthy cases that were hard for the agent policy to generalize to. 

\begin{figure}[b!]
  \centering
  \begin{subfigure}[b]{0.32\linewidth}
    \includegraphics[width=\linewidth]{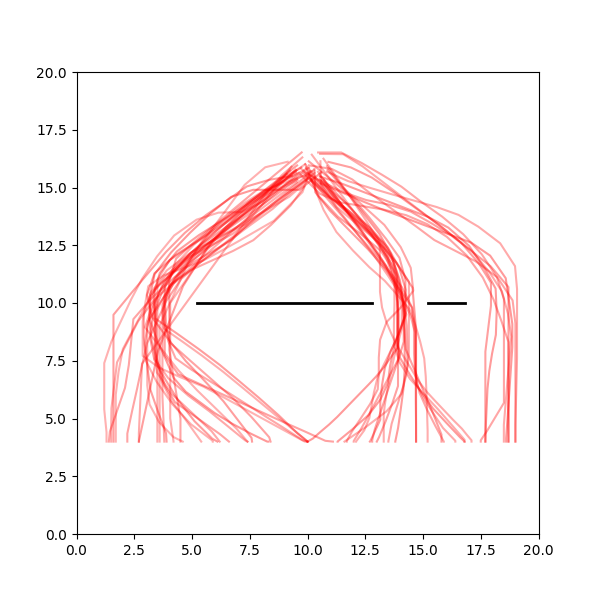}
     \caption{DDPG-LfD}
  \end{subfigure}
  \begin{subfigure}[b]{0.32\linewidth}
    \includegraphics[width=\linewidth]{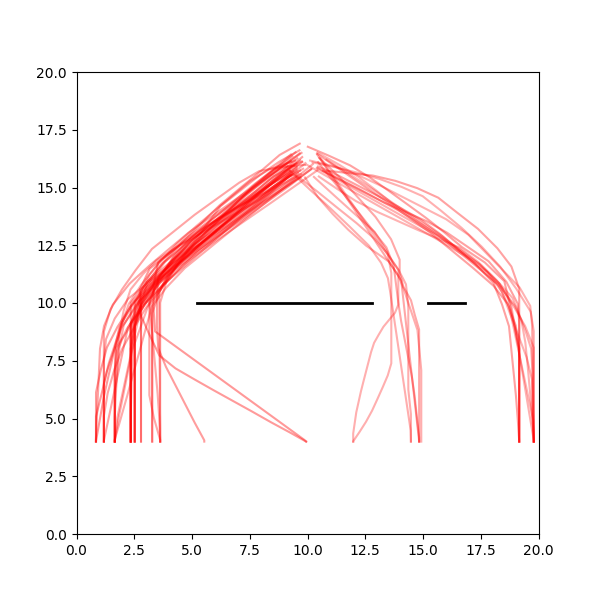}
    \caption{EARLY}
  \end{subfigure}
  \begin{subfigure}[b]{0.32\linewidth}
    \includegraphics[width=\linewidth]{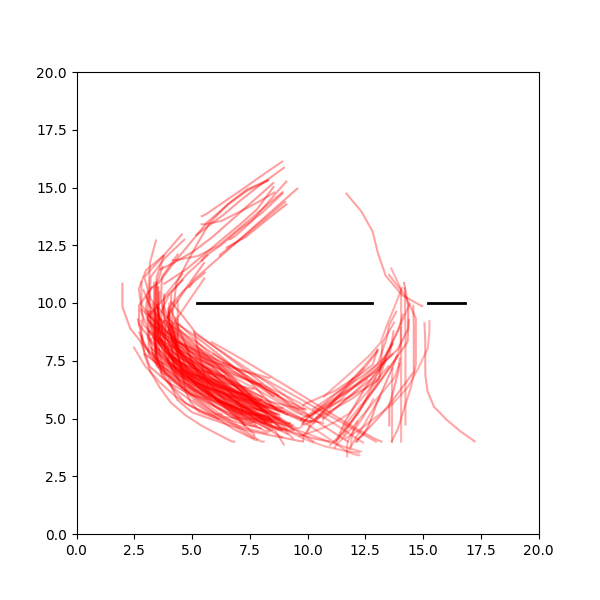}
    \caption{I-ARLD}
  \end{subfigure}
  
  \caption{Distribution of provided demonstrations from one of the human participants using different baseline methods.}
  \label{demo distributions}
\end{figure}

Indeed, as shown in Figure \ref{demo distributions}, what the agent needs most help with is highly different from what the human expert believed to be most helpful for agent learning. By contrast, our method accelerated the learning process by helping identify the cases that were most learning-beneficial, leading to faster convergence to the expert-level performance. Although I-ARLD also enabled the agent to ask for help when stuck in local optima, it spent most of its demonstration budget on showing the agent how to get out of the local optima, as opposed to how to avoid getting into the local optima in the first place, which leads to a slower converge compared with our method.

% Similarly, for the simulated navigation tasks with human expert demonstrations, we measured the average episode rewards and success rate over $1000$ randomly initialized test episodes at an interval of $1000$ environment steps, and presented the results that were averaged over all $5$ human experts. As shown in Figure ~\ref{reward and success plot for nav}, with the same amount of $60$ demonstrations whose feature values (i.e., initial states) were selected by human experts themselves, the resulting policy performance was even worse than our method. The produced policies did not converge to the expert policy within the given amount of environment steps and barely solved the tasks. It indicated that even if human experts could always provide optimal actions for all the states encountered in one single demonstration (i.e., optimal in executing the task), they may be sub-optimal or bad in selecting the best set of demonstrations that is most beneficial for policy learning (i.e., sub-optimal in teaching the task). Without active involvement from the learner side as in our method, such sub-optimality in teaching could even consequently lead to catastrophic training performance.

\subsubsection{User Experience}
To investigate the perceived task load of our method, we conducted a one-way repeated ANOVA test for each metric of the standard NASA-TLX questionnaire respectively. As shown in Figure \ref{user experience}, our method required lower average demands from human experts than the other two baselines in general. More specifically, there was a significant difference in mental demand among the three learning algorithms ($F(2, 34)=8.96, p<.01$) with a large effect size ($\eta^2=0.18$). The Tukey HSD post hoc test indicated that our method ($M=4.56, SD=2.64$) posed a significantly lower mental demand than both DDPG-LfD ($M=9.06, SD=5.53$) and I-ARLD ($M=8.33, SD=4.21$). However, there was no significant difference between DDPG-LfD and I-ARLD. For other metrics of perceived task load, although we did not observe any statistical significance because of the relatively small sample size, our method exhibited a smaller average demand than the other two baselines except for the temporal demand. This was reasonable considering that the human experts were able to choose their demonstrations at their own paces when using DDPG-LfD, while the learning agent would decide the timing of each query in both I-ARLD and EARLY. Despite this, our method was yet less temporally demanding than I-ARLD, indicating an improved temporal experience.

In addition to the perceived task load, we also conducted a one-way repeated ANOVA test for the total amount of human time spent by each method. As shown in Figure \ref{user experience}, there was a significant difference in the amount of human time among the three learning algorithms ($F(2, 34)=233.11, p<.001$) with a large effect size ($\eta^2=0.87$). According to the Tukey HSD post hoc test, we observed that our method ($M=3.22, SD=0.98$) consumed significantly less human time than DDPG-LfD ($M=7.07, SD=2.04$) and I-ARLD ($M=11.48, SD=0.44$), and DDPG-LfD consumed significantly less human time than I-ARLD. These results indicated that our method required less time effort from human experts, further validating the improved user experience of our method than the baselines.

\subsection{Limitations}
In this work, we chose the initial state $s_0$ as the feature $\varphi_i$ of an episodic roll-out trajectory $\xi_{\pi}^i$ under the policy $\pi$. This will make the probability distribution of feature $\varphi$ be independent from the current policy $\pi$ and only dependent on a stationary initial state distribution $\mu(s_0)$. In more general cases, the probability distribution of feature points will also be dependent on the current parametrized agent policy $\pi_{\phi}$ that is non-stationary during the training process. And if the policy updates along the wrong direction or gets stuck in a local optima that is worse than the expert policy, it may make the estimation of uncertainty distribution in the feature space far less accurate and constrain the exploration in the feature space, leading to queries wasted on areas that may not be much beneficial to accelerate policy learning.

Furthermore, when querying an episodic expert demonstration $\xi_{\pi_{demo}^k}$ whose feature value is expected to be $\varphi_k$, we assumed that the expert will always be able to provide a demonstration whose feature value is exactly equal to $\varphi_k$. In practice, especially in the cases of human experts, the feature $\varphi_{real}$ of the obtained expert demonstrations may follow an unknown distribution that is related to $\varphi_k$. Therefore, a more general query strategy should not only consider how uncertain the agent is about each individual feature points, but also take into account how possible it is to obtain an expert demonstration that is featured exactly on the uncertain feature point if the agent queries about it. 
%%%%%%%%%%%%%%%%%%%%%%%%%%%%%%%%%%%%%%%%%%%%%%%%%%%%%%%%%%%%%%%%%%%%%%%%

\section{Conclusions}

In this work, we present a framework that enables the agent to solve sequential decision-making problems by actively querying episodic demonstrations from the expert in a trajectory-based feature space. We constructed a trajectory-based measurement to evaluate the uncertainty of the agent policy and utilized it to determine the query timing and generate feature-oriented queries that may most influence the uncertainty distribution and consequently accelerate policy learning. By querying episodic demonstrations of target feature values, our method achieved better learning performance and improved the user experience of human demonstrators. We verified the effectiveness of our method in three simulated navigation tasks with scaling levels of difficulty with both oracle-simulated and human expert demonstrators. The results showed that our method maintained strong performance in all tasks and converged to the expert policy much faster than other baseline methods. Furthermore, our method achieved a better user experience in perceived task load while consuming significantly less human time. For future work, we plan to extend our method to more general feature designs, where the ongoing agent policy will also influence the probability distribution of feature points, and take into account the uncertainty that may be introduced by the discrepancy of the feature values between the obtained expert demonstrations and queried ones.

%%%%%%%%%%%%%%%%%%%%%%%%%%%%%%%%%%%%%%%%%%%%%%%%%%%%%%%%%%%%%%%%%%%%%%%%

%%
%% The acknowledgments section is defined using the "acks" environment
%% (and NOT an unnumbered section). This ensures the proper
%% identification of the section in the article metadata, and the
%% consistent spelling of the heading.
% \begin{acks}
% To Robert, for the bagels and explaining CMYK and color spaces.
% \end{acks}

%%
%% The next two lines define the bibliography style to be used, and
%% the bibliography file.
\bibliographystyle{ACM-Reference-Format}
\bibliography{sample}

%%% -*-BibTeX-*-
%%% Do NOT edit. File created by BibTeX with style
%%% ACM-Reference-Format-Journals [18-Jan-2012].

\begin{thebibliography}{31}

%%% ====================================================================
%%% NOTE TO THE USER: you can override these defaults by providing
%%% customized versions of any of these macros before the \bibliography
%%% command.  Each of them MUST provide its own final punctuation,
%%% except for \shownote{}, \showDOI{}, and \showURL{}.  The latter two
%%% do not use final punctuation, in order to avoid confusing it with
%%% the Web address.
%%%
%%% To suppress output of a particular field, define its macro to expand
%%% to an empty string, or better, \unskip, like this:
%%%
%%% \newcommand{\showDOI}[1]{\unskip}   % LaTeX syntax
%%%
%%% \def \showDOI #1{\unskip}           % plain TeX syntax
%%%
%%% ====================================================================

\ifx \showCODEN    \undefined \def \showCODEN     #1{\unskip}     \fi
\ifx \showDOI      \undefined \def \showDOI       #1{#1}\fi
\ifx \showISBNx    \undefined \def \showISBNx     #1{\unskip}     \fi
\ifx \showISBNxiii \undefined \def \showISBNxiii  #1{\unskip}     \fi
\ifx \showISSN     \undefined \def \showISSN      #1{\unskip}     \fi
\ifx \showLCCN     \undefined \def \showLCCN      #1{\unskip}     \fi
\ifx \shownote     \undefined \def \shownote      #1{#1}          \fi
\ifx \showarticletitle \undefined \def \showarticletitle #1{#1}   \fi
\ifx \showURL      \undefined \def \showURL       {\relax}        \fi
% The following commands are used for tagged output and should be
% invisible to TeX
\providecommand\bibfield[2]{#2}
\providecommand\bibinfo[2]{#2}
\providecommand\natexlab[1]{#1}
\providecommand\showeprint[2][]{arXiv:#2}

\bibitem[Cakmak and Thomaz(2011)]%
        {cakmak2011active}
\bibfield{author}{\bibinfo{person}{Maya Cakmak} {and} \bibinfo{person}{Andrea~L Thomaz}.} \bibinfo{year}{2011}\natexlab{}.
\newblock \showarticletitle{Active learning with mixed query types in learning from demonstration}. In \bibinfo{booktitle}{\emph{Proc. of the ICML workshop on new developments in imitation learning}}. Citeseer.
\newblock


\bibitem[Chen et~al\mbox{.}(2023)]%
        {chenactive}
\bibfield{author}{\bibinfo{person}{Ming-Hsin Chen}, \bibinfo{person}{Si-An Chen}, {and} \bibinfo{person}{Hsuan-Tien Lin}.} \bibinfo{year}{2023}\natexlab{}.
\newblock \showarticletitle{Active Reinforcement Learning from Demonstration in Continuous Action Spaces}. In \bibinfo{booktitle}{\emph{AI and HCI Workshop at the 40th International Conference on Machine Learning (ICML), Honolulu, Hawaii, USA. 2023}}.
\newblock


\bibitem[Chen et~al\mbox{.}(2020)]%
        {chen2020active}
\bibfield{author}{\bibinfo{person}{Si-An Chen}, \bibinfo{person}{Voot Tangkaratt}, \bibinfo{person}{Hsuan-Tien Lin}, {and} \bibinfo{person}{Masashi Sugiyama}.} \bibinfo{year}{2020}\natexlab{}.
\newblock \showarticletitle{Active deep Q-learning with demonstration}.
\newblock \bibinfo{journal}{\emph{Machine Learning}}  \bibinfo{volume}{109} (\bibinfo{year}{2020}), \bibinfo{pages}{1699--1725}.
\newblock


\bibitem[Chernova and Veloso(2009)]%
        {chernova2009interactive}
\bibfield{author}{\bibinfo{person}{Sonia Chernova} {and} \bibinfo{person}{Manuela Veloso}.} \bibinfo{year}{2009}\natexlab{}.
\newblock \showarticletitle{Interactive policy learning through confidence-based autonomy}.
\newblock \bibinfo{journal}{\emph{Journal of Artificial Intelligence Research}}  \bibinfo{volume}{34} (\bibinfo{year}{2009}), \bibinfo{pages}{1--25}.
\newblock


\bibitem[Gehring and Precup(2013)]%
        {gehring2013smart}
\bibfield{author}{\bibinfo{person}{Clement Gehring} {and} \bibinfo{person}{Doina Precup}.} \bibinfo{year}{2013}\natexlab{}.
\newblock \showarticletitle{Smart exploration in reinforcement learning using absolute temporal difference errors}. In \bibinfo{booktitle}{\emph{Proceedings of the 2013 international conference on Autonomous agents and multi-agent systems}}. \bibinfo{pages}{1037--1044}.
\newblock


\bibitem[Gleave et~al\mbox{.}(2022)]%
        {gleave2022imitation}
\bibfield{author}{\bibinfo{person}{Adam Gleave}, \bibinfo{person}{Mohammad Taufeeque}, \bibinfo{person}{Juan Rocamonde}, \bibinfo{person}{Erik Jenner}, \bibinfo{person}{Steven~H. Wang}, \bibinfo{person}{Sam Toyer}, \bibinfo{person}{Maximilian Ernestus}, \bibinfo{person}{Nora Belrose}, \bibinfo{person}{Scott Emmons}, {and} \bibinfo{person}{Stuart Russell}.} \bibinfo{year}{2022}\natexlab{}.
\newblock \bibinfo{title}{imitation: Clean Imitation Learning Implementations}.
\newblock \bibinfo{howpublished}{arXiv:2211.11972v1 [cs.LG]}.
\newblock
\showeprint[arxiv]{2211.11972}~[cs.LG]
\urldef\tempurl%
\url{https://arxiv.org/abs/2211.11972}
\showURL{%
\tempurl}


\bibitem[Haarnoja et~al\mbox{.}(2018)]%
        {haarnoja2018soft}
\bibfield{author}{\bibinfo{person}{Tuomas Haarnoja}, \bibinfo{person}{Aurick Zhou}, \bibinfo{person}{Pieter Abbeel}, {and} \bibinfo{person}{Sergey Levine}.} \bibinfo{year}{2018}\natexlab{}.
\newblock \showarticletitle{Soft actor-critic: Off-policy maximum entropy deep reinforcement learning with a stochastic actor}. In \bibinfo{booktitle}{\emph{International conference on machine learning}}. PMLR, \bibinfo{pages}{1861--1870}.
\newblock


\bibitem[Hawke et~al\mbox{.}(2020)]%
        {hawke2020urban}
\bibfield{author}{\bibinfo{person}{Jeffrey Hawke}, \bibinfo{person}{Richard Shen}, \bibinfo{person}{Corina Gurau}, \bibinfo{person}{Siddharth Sharma}, \bibinfo{person}{Daniele Reda}, \bibinfo{person}{Nikolay Nikolov}, \bibinfo{person}{Przemys{\l}aw Mazur}, \bibinfo{person}{Sean Micklethwaite}, \bibinfo{person}{Nicolas Griffiths}, \bibinfo{person}{Amar Shah}, {et~al\mbox{.}}} \bibinfo{year}{2020}\natexlab{}.
\newblock \showarticletitle{Urban driving with conditional imitation learning}. In \bibinfo{booktitle}{\emph{2020 IEEE International Conference on Robotics and Automation (ICRA)}}. IEEE, \bibinfo{pages}{251--257}.
\newblock


\bibitem[Hester et~al\mbox{.}(2018)]%
        {hester2018deep}
\bibfield{author}{\bibinfo{person}{Todd Hester}, \bibinfo{person}{Matej Vecerik}, \bibinfo{person}{Olivier Pietquin}, \bibinfo{person}{Marc Lanctot}, \bibinfo{person}{Tom Schaul}, \bibinfo{person}{Bilal Piot}, \bibinfo{person}{Dan Horgan}, \bibinfo{person}{John Quan}, \bibinfo{person}{Andrew Sendonaris}, \bibinfo{person}{Ian Osband}, {et~al\mbox{.}}} \bibinfo{year}{2018}\natexlab{}.
\newblock \showarticletitle{Deep q-learning from demonstrations}. In \bibinfo{booktitle}{\emph{Proceedings of the AAAI conference on artificial intelligence}}, Vol.~\bibinfo{volume}{32}.
\newblock


\bibitem[Ho and Ermon(2016)]%
        {ho2016generative}
\bibfield{author}{\bibinfo{person}{Jonathan Ho} {and} \bibinfo{person}{Stefano Ermon}.} \bibinfo{year}{2016}\natexlab{}.
\newblock \showarticletitle{Generative adversarial imitation learning}.
\newblock \bibinfo{journal}{\emph{Advances in neural information processing systems}}  \bibinfo{volume}{29} (\bibinfo{year}{2016}).
\newblock


\bibitem[Hou et~al\mbox{.}(2023)]%
        {hou2023shaping}
\bibfield{author}{\bibinfo{person}{Muhan Hou}, \bibinfo{person}{Koen Hindriks}, \bibinfo{person}{A.E. Eiben}, {and} \bibinfo{person}{Kim Baraka}.} \bibinfo{year}{2023}\natexlab{}.
\newblock \showarticletitle{Shaping Imbalance into Balance: Active Robot Guidance of Human Teachers for Better Learning from Demonstrations}. In \bibinfo{booktitle}{\emph{2023 31st IEEE International Conference on Robot and Human Interactive Communication (RO-MAN)}}. IEEE.
\newblock


\bibitem[Johns(2022)]%
        {johns2022back}
\bibfield{author}{\bibinfo{person}{Edward Johns}.} \bibinfo{year}{2022}\natexlab{}.
\newblock \showarticletitle{Back to reality for imitation learning}. In \bibinfo{booktitle}{\emph{Conference on Robot Learning}}. PMLR, \bibinfo{pages}{1764--1768}.
\newblock


\bibitem[Judah et~al\mbox{.}(2014)]%
        {judah2014active}
\bibfield{author}{\bibinfo{person}{Kshitij Judah}, \bibinfo{person}{Alan~Paul Fern}, \bibinfo{person}{Thomas~G Dietterich}, {and} \bibinfo{person}{Prasad Tadepalli}.} \bibinfo{year}{2014}\natexlab{}.
\newblock \showarticletitle{Active lmitation learning: formal and practical reductions to IID learning.}
\newblock \bibinfo{journal}{\emph{J. Mach. Learn. Res.}} \bibinfo{volume}{15}, \bibinfo{number}{1} (\bibinfo{year}{2014}), \bibinfo{pages}{3925--3963}.
\newblock


\bibitem[Kang et~al\mbox{.}(2018)]%
        {kang2018policy}
\bibfield{author}{\bibinfo{person}{Bingyi Kang}, \bibinfo{person}{Zequn Jie}, {and} \bibinfo{person}{Jiashi Feng}.} \bibinfo{year}{2018}\natexlab{}.
\newblock \showarticletitle{Policy optimization with demonstrations}. In \bibinfo{booktitle}{\emph{International conference on machine learning}}. PMLR, \bibinfo{pages}{2469--2478}.
\newblock


\bibitem[Karaman and Frazzoli(2010)]%
        {karaman2010incremental}
\bibfield{author}{\bibinfo{person}{Sertac Karaman} {and} \bibinfo{person}{Emilio Frazzoli}.} \bibinfo{year}{2010}\natexlab{}.
\newblock \showarticletitle{Incremental sampling-based algorithms for optimal motion planning}.
\newblock \bibinfo{journal}{\emph{Robotics Science and Systems VI}} \bibinfo{volume}{104}, \bibinfo{number}{2} (\bibinfo{year}{2010}), \bibinfo{pages}{267--274}.
\newblock


\bibitem[Kelly et~al\mbox{.}(2019)]%
        {kelly2019hg}
\bibfield{author}{\bibinfo{person}{Michael Kelly}, \bibinfo{person}{Chelsea Sidrane}, \bibinfo{person}{Katherine Driggs-Campbell}, {and} \bibinfo{person}{Mykel~J Kochenderfer}.} \bibinfo{year}{2019}\natexlab{}.
\newblock \showarticletitle{Hg-dagger: Interactive imitation learning with human experts}. In \bibinfo{booktitle}{\emph{2019 International Conference on Robotics and Automation (ICRA)}}. IEEE, \bibinfo{pages}{8077--8083}.
\newblock


\bibitem[Liu et~al\mbox{.}(2022)]%
        {liu2022improved}
\bibfield{author}{\bibinfo{person}{Haochen Liu}, \bibinfo{person}{Zhiyu Huang}, \bibinfo{person}{Jingda Wu}, {and} \bibinfo{person}{Chen Lv}.} \bibinfo{year}{2022}\natexlab{}.
\newblock \showarticletitle{Improved deep reinforcement learning with expert demonstrations for urban autonomous driving}. In \bibinfo{booktitle}{\emph{2022 IEEE Intelligent Vehicles Symposium (IV)}}. IEEE, \bibinfo{pages}{921--928}.
\newblock


\bibitem[Mnih et~al\mbox{.}(2013)]%
        {mnih2013playing}
\bibfield{author}{\bibinfo{person}{Volodymyr Mnih}, \bibinfo{person}{Koray Kavukcuoglu}, \bibinfo{person}{David Silver}, \bibinfo{person}{Alex Graves}, \bibinfo{person}{Ioannis Antonoglou}, \bibinfo{person}{Daan Wierstra}, {and} \bibinfo{person}{Martin Riedmiller}.} \bibinfo{year}{2013}\natexlab{}.
\newblock \showarticletitle{Playing atari with deep reinforcement learning}.
\newblock \bibinfo{journal}{\emph{arXiv preprint arXiv:1312.5602}} (\bibinfo{year}{2013}).
\newblock


\bibitem[Nair et~al\mbox{.}(2020)]%
        {nair2020awac}
\bibfield{author}{\bibinfo{person}{Ashvin Nair}, \bibinfo{person}{Abhishek Gupta}, \bibinfo{person}{Murtaza Dalal}, {and} \bibinfo{person}{Sergey Levine}.} \bibinfo{year}{2020}\natexlab{}.
\newblock \showarticletitle{Awac: Accelerating online reinforcement learning with offline datasets}.
\newblock \bibinfo{journal}{\emph{arXiv preprint arXiv:2006.09359}} (\bibinfo{year}{2020}).
\newblock


\bibitem[Nair et~al\mbox{.}(2018)]%
        {nair2018overcoming}
\bibfield{author}{\bibinfo{person}{Ashvin Nair}, \bibinfo{person}{Bob McGrew}, \bibinfo{person}{Marcin Andrychowicz}, \bibinfo{person}{Wojciech Zaremba}, {and} \bibinfo{person}{Pieter Abbeel}.} \bibinfo{year}{2018}\natexlab{}.
\newblock \showarticletitle{Overcoming exploration in reinforcement learning with demonstrations}. In \bibinfo{booktitle}{\emph{2018 IEEE international conference on robotics and automation (ICRA)}}. IEEE, \bibinfo{pages}{6292--6299}.
\newblock


\bibitem[Packard and Ontan{\'o}n(2017)]%
        {packard2017policies}
\bibfield{author}{\bibinfo{person}{Brandon Packard} {and} \bibinfo{person}{Santiago Ontan{\'o}n}.} \bibinfo{year}{2017}\natexlab{}.
\newblock \showarticletitle{Policies for active learning from demonstration}. In \bibinfo{booktitle}{\emph{2017 AAAI Spring Symposium Series}}.
\newblock


\bibitem[Pomerleau(1988)]%
        {pomerleau1988alvinn}
\bibfield{author}{\bibinfo{person}{Dean~A Pomerleau}.} \bibinfo{year}{1988}\natexlab{}.
\newblock \showarticletitle{Alvinn: An autonomous land vehicle in a neural network}.
\newblock \bibinfo{journal}{\emph{Advances in neural information processing systems}}  \bibinfo{volume}{1} (\bibinfo{year}{1988}).
\newblock


\bibitem[Ram{\'\i}rez et~al\mbox{.}(2022)]%
        {ramirez2022model}
\bibfield{author}{\bibinfo{person}{Jorge Ram{\'\i}rez}, \bibinfo{person}{Wen Yu}, {and} \bibinfo{person}{Adolfo Perrusqu{\'\i}a}.} \bibinfo{year}{2022}\natexlab{}.
\newblock \showarticletitle{Model-free reinforcement learning from expert demonstrations: a survey}.
\newblock \bibinfo{journal}{\emph{Artificial Intelligence Review}} (\bibinfo{year}{2022}), \bibinfo{pages}{1--29}.
\newblock


\bibitem[Rigter et~al\mbox{.}(2020)]%
        {rigter2020framework}
\bibfield{author}{\bibinfo{person}{Marc Rigter}, \bibinfo{person}{Bruno Lacerda}, {and} \bibinfo{person}{Nick Hawes}.} \bibinfo{year}{2020}\natexlab{}.
\newblock \showarticletitle{A framework for learning from demonstration with minimal human effort}.
\newblock \bibinfo{journal}{\emph{IEEE Robotics and Automation Letters}} \bibinfo{volume}{5}, \bibinfo{number}{2} (\bibinfo{year}{2020}), \bibinfo{pages}{2023--2030}.
\newblock


\bibitem[Schaul et~al\mbox{.}(2015)]%
        {schaul2015prioritized}
\bibfield{author}{\bibinfo{person}{Tom Schaul}, \bibinfo{person}{John Quan}, \bibinfo{person}{Ioannis Antonoglou}, {and} \bibinfo{person}{David Silver}.} \bibinfo{year}{2015}\natexlab{}.
\newblock \showarticletitle{Prioritized experience replay}.
\newblock \bibinfo{journal}{\emph{arXiv preprint arXiv:1511.05952}} (\bibinfo{year}{2015}).
\newblock


\bibitem[Silver et~al\mbox{.}(2012)]%
        {silver2012active}
\bibfield{author}{\bibinfo{person}{David Silver}, \bibinfo{person}{J~Andrew Bagnell}, {and} \bibinfo{person}{Anthony Stentz}.} \bibinfo{year}{2012}\natexlab{}.
\newblock \showarticletitle{Active learning from demonstration for robust autonomous navigation}. In \bibinfo{booktitle}{\emph{2012 IEEE International Conference on Robotics and Automation}}. IEEE, \bibinfo{pages}{200--207}.
\newblock


\bibitem[Singh et~al\mbox{.}(2020)]%
        {singh2020parrot}
\bibfield{author}{\bibinfo{person}{Avi Singh}, \bibinfo{person}{Huihan Liu}, \bibinfo{person}{Gaoyue Zhou}, \bibinfo{person}{Albert Yu}, \bibinfo{person}{Nicholas Rhinehart}, {and} \bibinfo{person}{Sergey Levine}.} \bibinfo{year}{2020}\natexlab{}.
\newblock \showarticletitle{Parrot: Data-driven behavioral priors for reinforcement learning}.
\newblock \bibinfo{journal}{\emph{arXiv preprint arXiv:2011.10024}} (\bibinfo{year}{2020}).
\newblock


\bibitem[Subramanian et~al\mbox{.}(2016)]%
        {subramanian2016exploration}
\bibfield{author}{\bibinfo{person}{Kaushik Subramanian}, \bibinfo{person}{Charles~L Isbell~Jr}, {and} \bibinfo{person}{Andrea~L Thomaz}.} \bibinfo{year}{2016}\natexlab{}.
\newblock \showarticletitle{Exploration from demonstration for interactive reinforcement learning}. In \bibinfo{booktitle}{\emph{Proceedings of the 2016 international conference on autonomous agents \& multiagent systems}}. \bibinfo{pages}{447--456}.
\newblock


\bibitem[Taylor et~al\mbox{.}(2011)]%
        {taylor2011integrating}
\bibfield{author}{\bibinfo{person}{Matthew~E Taylor}, \bibinfo{person}{Halit~Bener Suay}, {and} \bibinfo{person}{Sonia Chernova}.} \bibinfo{year}{2011}\natexlab{}.
\newblock \showarticletitle{Integrating reinforcement learning with human demonstrations of varying ability}. In \bibinfo{booktitle}{\emph{The 10th International Conference on Autonomous Agents and Multiagent Systems-Volume 2}}. \bibinfo{pages}{617--624}.
\newblock


\bibitem[Vecerik et~al\mbox{.}(2017)]%
        {vecerik2017leveraging}
\bibfield{author}{\bibinfo{person}{Mel Vecerik}, \bibinfo{person}{Todd Hester}, \bibinfo{person}{Jonathan Scholz}, \bibinfo{person}{Fumin Wang}, \bibinfo{person}{Olivier Pietquin}, \bibinfo{person}{Bilal Piot}, \bibinfo{person}{Nicolas Heess}, \bibinfo{person}{Thomas Roth{\"o}rl}, \bibinfo{person}{Thomas Lampe}, {and} \bibinfo{person}{Martin Riedmiller}.} \bibinfo{year}{2017}\natexlab{}.
\newblock \showarticletitle{Leveraging demonstrations for deep reinforcement learning on robotics problems with sparse rewards}.
\newblock \bibinfo{journal}{\emph{arXiv preprint arXiv:1707.08817}} (\bibinfo{year}{2017}).
\newblock


\bibitem[Wang and Taylor(2017)]%
        {wang2017improving}
\bibfield{author}{\bibinfo{person}{Zhaodong Wang} {and} \bibinfo{person}{Matthew~E Taylor}.} \bibinfo{year}{2017}\natexlab{}.
\newblock \showarticletitle{Improving Reinforcement Learning with Confidence-Based Demonstrations.}. In \bibinfo{booktitle}{\emph{IJCAI}}. \bibinfo{pages}{3027--3033}.
\newblock


\end{thebibliography}

\end{document}